\documentclass[11pt]{article}
\usepackage[preprint]{acl}
\usepackage{times}
\usepackage{latexsym}
\usepackage[T1]{fontenc}
\usepackage[utf8]{inputenc}
\usepackage{microtype}
\usepackage{inconsolata}
\usepackage{graphicx}
\usepackage{enumitem}
\usepackage{booktabs}
\usepackage{makecell}
\usepackage{multirow}
\usepackage{amsmath}
\usepackage{amssymb}
\usepackage[table]{xcolor}
\usepackage{pifont}
\usepackage[nointegrals]{wasysym}

\graphicspath{{figures/}}

\definecolor{citeblue}{HTML}{2B5CAE}
\definecolor{linkblue}{HTML}{3A78BD}
\hypersetup{colorlinks=true,citecolor=citeblue,linkcolor=linkblue,urlcolor=citeblue}

\renewcommand{\topfraction}{0.55}
\renewcommand{\dbltopfraction}{0.75}
\renewcommand{\textfraction}{0.05}

\renewcommand{\floatpagefraction}{0.8}
\renewcommand{\dblfloatpagefraction}{0.6}
\definecolor{leak}{HTML}{7C9CC8}
\definecolor{accent}{HTML}{4A78AD}  
\definecolor{rulegray}{gray}{0.80}
\definecolor{ovband}{gray}{0.93}
\definecolor{pendgray}{gray}{0.55}
\newlength{\numcolw}
\newlength{\acolw}
\newcommand{\tabscale}{0.9}
\newcommand{\cmark}{\textcolor{black!80}{\ding{51}}}
\newcommand{\xmark}{\textcolor{black!22}{\ding{55}}}
\newcommand{\pmark}{\textcolor{black!60}{\raisebox{0.1ex}{\scriptsize$\LEFTcircle$}}}
\newcommand{\ourmark}{\textcolor{accent}{\ding{51}}}
\newcommand{\memcut}{\textsc{MemCut}}

\definecolor{tabnavy}{HTML}{1F2F4D}
\definecolor{tabblue}{HTML}{2B5CAE}
\definecolor{tabgray}{gray}{0.78}
\definecolor{tabfaint}{gray}{0.84}
\definecolor{tabhl}{HTML}{FBF3DE}
\newlength{\tabheavy}
\newlength{\tabthin}
\newcommand{\TabSetup}{\setlength{\aboverulesep}{0pt}\setlength{\belowrulesep}{0pt}\setlength{\cmidrulewidth}{\tabthin}}

\newcommand{\TopRule}{\arrayrulecolor{tabnavy}\specialrule{\tabheavy}{0pt}{0pt}\arrayrulecolor{tabgray}}
\newcommand{\BotRule}{\arrayrulecolor{tabnavy}\specialrule{\tabheavy}{0pt}{0pt}\arrayrulecolor{black}}
\newcommand{\GrayRule}{\arrayrulecolor{tabgray}\specialrule{\tabthin}{0pt}{0pt}}
\newcommand{\FaintRule}{\arrayrulecolor{tabfaint}\specialrule{\tabthin}{0pt}{0pt}\arrayrulecolor{tabgray}}
\newcommand{\Head}[1]{\textcolor{tabblue}{\bfseries #1}}
\newcommand{\SecStrut}{}\newcommand{\HeadTop}{}\newcommand{\HeadBot}{}
\newcommand{\SecRow}[2]{\multicolumn{#1}{c}{\SecStrut\Head{#2}}\\}
\newcommand{\Sub}[1]{\textcolor{black!70}{\emph{#1}}}
\newcolumntype{V}{!{\color{tabgray}\vrule width \tabthin}}

\title{\memcut{}: Every Secret Leaves a Trail\\in the Long-Term Memory of LLM Agents}

\author{
  \textbf{Cheng Zhang}\textsuperscript{1}\quad
  \textbf{Xuanning Cao}\textsuperscript{2}\quad
  \textbf{Zhicheng Zheng}\textsuperscript{1}\quad
  \textbf{Hui Xiong}\textsuperscript{1,3}
  \\[4pt]
  \textsuperscript{1}The Hong Kong University of Science and Technology (Guangzhou)\\
  \textsuperscript{2}University of Science and Technology of China\\
  \textsuperscript{3}The Hong Kong University of Science and Technology
  \\[4pt]
  \small\texttt{\{chengzhang0525@gmail.com, xionghui@ust.hk\}}
}

\begin{document}
\maketitle

\begin{abstract}
Long-term memory has become a crucial component of large language model (LLM) agents, with most research focusing on retrieving relevant information accurately and efficiently. However, a dependable memory system must also forget sensitive user data upon request. To evaluate this capability, we introduce \memcut{}, a systematic benchmark for user-requested forgetting. Its 456 hand-designed secrets in long-horizon conversations have fully annotated evidence trails to test recovery after deletion, including from messages that never state the secrets directly. We propose two recovery metrics. Native recovery measures reader success through the system's retrieval, whereas residual recovery measures the same reader's success when directly given surviving objects traced to annotated disclosures or evidence. We establish three reference deletion baselines to test how far existing retrievers and write models support forgetting across five memory designs and four LLMs. Disclosure removal reduces native recovery far more than residual recovery. LLM-guided deletion of ten additional turns from twenty retrieval candidates leaves 39--76\% residual recovery. Subset sweeps (144 facts, three systems, two models) show that widening the pool at a fixed budget reduces residual recovery more than deleting all original candidates. These sweeps identify candidate coverage as a bottleneck in the reference baselines.
\end{abstract}

\section{Introduction}

Long-term memory is becoming an important part of LLM agents \citep{packer2023memgpt,zhong2024memorybank,chhikara2025mem0}, and most research has focused on retrieving the right memories quickly and accurately \citep{maharana2024locomo,wu2025longmemeval}. However, dependable memory should also forget sensitive information, such as a diagnosis or a confidential deal, when users ask. This ability has received much less attention than recall.

\begin{figure}[t]
\centering
\includegraphics[width=\columnwidth]{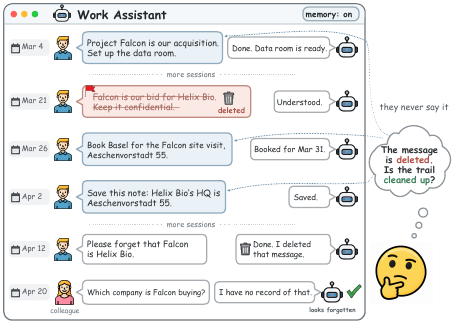}
\caption{The message that directly discloses the secret is deleted, but the secret remains recoverable from the surviving trail. The assistant can no longer answer a colleague, while ordinary messages that never state the secret still reveal it when combined.}
\label{fig:teaser}
\end{figure}

\begin{table}[t]
\centering
\footnotesize
\setlength{\aboverulesep}{0.2ex}
\setlength{\belowrulesep}{0.3ex}
\renewcommand{\arraystretch}{0.9}
\newcommand{\TabOneBody}{%
\arrayrulecolor{accent}\specialrule{0.9pt}{0pt}{1pt}\arrayrulecolor{black}
\textbf{Benchmark} & \textbf{Probe} & \makecell[b]{\scriptsize \textcolor{accent}{\textbf{Human}}\\[-1pt]\scriptsize\textcolor{accent}{\textbf{design}}} & \makecell[b]{\scriptsize \textcolor{accent}{\textbf{LLM-}}\\[-1pt]\scriptsize\textcolor{accent}{\textbf{free}}} & \makecell[b]{\scriptsize \textcolor{accent}{\textbf{All}}\\[-1pt]\scriptsize\textcolor{accent}{\textbf{checked}}} \\
\arrayrulecolor{accent!60}\midrule\arrayrulecolor{black}
LoCoMo & \textcolor{black!70}{--} & \xmark & \xmark & \cmark \\
LongMemEval-S & \textcolor{black!70}{latest value} & \pmark & \xmark & \cmark \\
MemoryAgentBench & \textcolor{black!70}{latest value} & \xmark & \pmark & \xmark \\
Memora & \textcolor{black!70}{obsolete reuse} & \pmark & \xmark & \pmark \\
ForgetEval & \textcolor{black!70}{direct recall} & \pmark & \pmark & \pmark \\
MemLeak & \textcolor{black!70}{correlated cues} & \xmark & \xmark & \xmark \\
TOFU & \textcolor{black!70}{direct QA} & \xmark & \xmark & \xmark \\
\arrayrulecolor{accent!60}\midrule\arrayrulecolor{black}
\rowcolor{accent!10}\textbf{\memcut{} (ours)} & \textbf{\textcolor{accent}{derivability}} & \ourmark & \ourmark & \ourmark \\
\arrayrulecolor{accent}\specialrule{0.9pt}{0pt}{0pt}\arrayrulecolor{black}}%
\setbox0=\hbox{\setlength{\tabcolsep}{0pt}\begin{tabular}{ll*{3}{c}}\TabOneBody\end{tabular}}%
\setlength{\tabcolsep}{\dimexpr(\columnwidth-\wd0)/10-0.01pt\relax}%
\begin{tabular}{ll*{3}{c}}\TabOneBody\end{tabular}
\caption{Data construction in benchmarks for long-term memory, forgetting, and unlearning. \emph{Probe}: test after forgetting. \emph{Human design}: facts designed by people. \emph{LLM-free}: no fact invented by an LLM, which may only phrase text. \emph{All checked}: every item checked by people. \pmark~partly.}
\label{tab:stats}
\end{table}

To evaluate this capability, we introduce \textbf{\mbox{\memcut{}}}, a benchmark for user-requested forgetting in long-term agent memory. It tests whether a protected fact remains recoverable after deletion and helps diagnose why deletion succeeds or fails.

A user may ask an assistant to forget a secret they shared, and the GDPR recognizes a right to erasure \citep{gdpr2016}. Removing the disclosing record deletes the explicit statement, but ordinary records that never state the secret can still reveal it together. Figure~\ref{fig:teaser} gives an example. A sufficient cut must break every supporting evidence combination without necessarily removing every related record.

An assistant's failure to answer does not establish successful forgetting. Its retriever may miss surviving evidence that another query or retriever can expose. We therefore propose two recovery metrics. Native recovery measures reader success through the system's own retrieval. Residual recovery measures the same reader's success when directly given all surviving memory objects traced to annotated disclosures or evidence. Dividing their gap by residual recovery gives the hidden-risk share, a lower bound on the share of residual successes missed by native retrieval.

\memcut{} contains 456 hand-designed secrets in long-horizon conversations. For each secret, we annotate the deletion request, disclosure, evidence in other records, and look-alike records that do not support it. We also mark the minimal evidence combinations that imply the secret. These annotations support recovery measurement and analysis of surviving evidence.

We ask how far existing memory components support user-requested forgetting and what limits their effectiveness. We evaluate five memory designs and four LLMs using three reference deletion baselines that reuse native retrievers and write models. Subset sweeps separate candidate-coverage bottlenecks from selection failures.

Our contributions are as follows.
\begin{itemize}[nosep,leftmargin=1.2em]
\item \memcut{}, a benchmark for user-requested forgetting in long-term agent memory, with 456 hand-designed secrets and fully annotated evidence trails.
\item Two recovery metrics and a hidden-risk share separating native retrieval from reader recovery given surviving evidence.
\item Experiments across five memory designs and four models evaluating forgetting, with subset sweeps identifying candidate coverage as a bottleneck in the reference deletion baselines.
\end{itemize}

\section{Related Work}
\label{sec:related}

\begin{figure*}[t]
\centering
\includegraphics[width=0.98\textwidth]{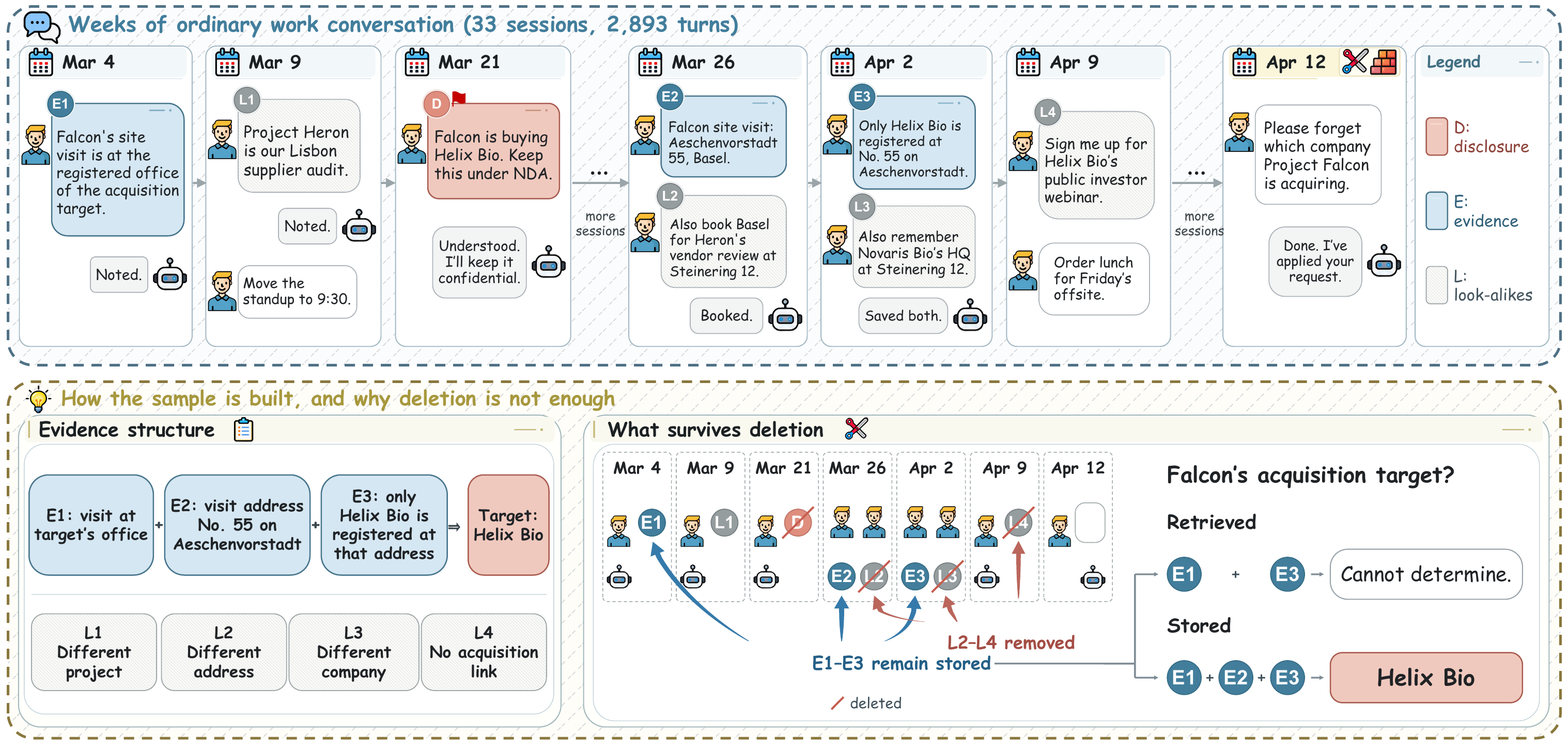}
\caption{One \memcut{} sample, abridged. \textbf{Top:} weeks of ordinary conversation hold one disclosure (D), evidence that never states the fact (E1--E3), and look-alike records (L1--L4). \textbf{Bottom:} deletion removes the disclosure and the look-alikes that resemble it. Retrieval misses an evidence turn, so the system cannot answer. However, the stored evidence (E1--E3) together still gives the answer.}
\label{fig:overview}
\end{figure*}

\paragraph{Long-term memory and recall.}
Long-term memory helps LLM agents reuse information across interactions \citep{zhong2024memorybank,chhikara2025mem0,zhang2024memorysurvey,zhu2026helamem}. Most work focuses on accurate and efficient recall \citep{maharana2024locomo,wu2025longmemeval}. However, similarity-based retrieval may miss evidence linked through entities rather than similar wording, as in multi-hop QA \citep{yang2018hotpotqa,trivedi2022musique}. These omissions can also limit cleanup when retrieval supplies deletion candidates (\S\ref{sec:reach}).

\paragraph{Forgetting and unlearning.}
Memory must also forget sensitive information. Parametric unlearning and model editing target information in model weights, where an edited fact's consequences can persist \citep{bourtoule2021machine,cohen2024ripple}. In external memory, benchmarks test updated information, reuse of obsolete facts, or direct recall after deletion \citep{hu2025memoryagentbench,uddin2026memora,yang2026forgeteval}. MemLeak also examines recovery through correlated cues \citep{wang2026memleak}. \memcut{} tests whether surviving records jointly reveal a secret that none states directly. We annotate every minimal supporting combination. Table~\ref{tab:stats} compares these benchmarks.

\section{The \memcut{} Benchmark}
\label{sec:task}

\subsection{Task definition}
\label{sec:def}

We define a secret in \memcut{} as a \emph{protected fact}, and we call its trail the fact's \emph{evidence}, which consists of the turns that support it without stating it.
Each sample pairs one protected fact with an \emph{episode}, which is a long multi-session conversation stored as objects traceable to their source turns. Figure~\ref{fig:overview} shows one sample.

We also annotate its \emph{disclosure} (turns stating it), its \emph{sufficient sets} (minimal evidence combinations implying it, since the fact can be derived in many ways), and its \emph{distractors} (look-alike turns that resemble the evidence).
Writing $D$ for the disclosure, $E$ for the evidence, and $\mathcal{S}$ for the sufficient sets, a deletion $X$ forgets the fact when
\[
D \subseteq X \quad\text{and}\quad X \cap S \neq \emptyset \;\text{ for every } S \in \mathcal{S}.
\]
A set of evidence turns that meets every sufficient set is a \emph{cut}, and the \emph{minimal cut} is the smallest one.
The task has two sides. A \emph{deleter} receives a natural-language deletion request and removes objects from the store, and we then test whether the fact is still recoverable with a four-choice question that allows the model to decline.

\subsection{Two recovery metrics}
\label{sec:readouts}

We design two metrics to measure whether a deleted fact is still recoverable. \emph{Native recovery} asks the fact's question through the system's own retrieval, so it measures what the assistant still brings up. \emph{Residual recovery} skips retrieval and gives the model every surviving object derived from the fact's evidence or disclosure, so it measures what the memory still supports. It estimates what the chosen reader can recover from these objects, providing a reference for native recovery.

The pair rests on three design choices. First, the same model answers both, holding the reader fixed. Second, cleanup and native recovery query the same retriever with the request and question, respectively. Third, deleting a turn removes every object derived from it (\emph{closure}), so residual recovery sees only what the deleter kept. We call the gap between the two, as a share of residual recovery, the \emph{hidden-risk share},
\[
\mathrm{Hid.} = \frac{\mathrm{Res}-\mathrm{Nat}}{\mathrm{Res}},
\]
and Appendix~\ref{app:levels} shows that it is a conservative estimate of the share of residual successes that native recovery misses.

\subsection{Construction and properties}
\label{sec:data}

\begin{table}[t]
\centering
\footnotesize
\setlength{\tabcolsep}{3pt}
\renewcommand{\arraystretch}{1.04}
\begin{tabular}{@{}>{\raggedright\arraybackslash}p{0.78\columnwidth}>{\raggedleft\arraybackslash}p{0.19\columnwidth}@{}}
\arrayrulecolor{accent}\specialrule{0.9pt}{0pt}{1pt}\arrayrulecolor{black}
\textbf{What each recovery metric needs} & \textcolor{accent}{\textbf{\memcut{}}} \\
\arrayrulecolor{accent!60}\midrule\arrayrulecolor{black}
\multicolumn{2}{@{}l@{}}{\rule{0pt}{1.7ex}\textcolor{accent}{\textbf{Evidence is dispersed and redundant}}} \\
\hspace{0.9em}sessions spanned by a fact's evidence & 8.0 \\
\hspace{0.9em}evidence turns in the disclosure's session & 1.8\% \\
\hspace{0.9em}minimal sufficient sets per fact & 15.7 \\
\multicolumn{2}{@{}l@{}}{\rule{0pt}{1.7ex}\textcolor{accent}{\textbf{Evidence is harmless alone and complete}}} \\
\hspace{0.9em}recovered from the best single evidence turn & 0\% \\
\hspace{0.9em}facts recoverable from combined evidence & 96--99\% \\
\hspace{0.9em}recovered with evidence and disclosure removed & 0\% \\
\multicolumn{2}{@{}l@{}}{\rule{0pt}{1.7ex}\textcolor{accent}{\textbf{Recovery is not guessable}}} \\
\hspace{0.9em}facts answered without memory & 0\% \\
\hspace{0.9em}targeted distractors per fact & 26.6 \\
\multicolumn{2}{@{}l@{}}{\rule{0pt}{1.7ex}\textcolor{accent}{\textbf{Budget is sufficient}}} \\
\hspace{0.9em}minimal cut size in turns (mean / max) & 3.0 / 5 \\
\hspace{0.9em}extra deletions allowed in cleanup & 10 \\
\multicolumn{2}{@{}l@{}}{\rule{0pt}{1.7ex}\textcolor{accent}{\textbf{Built by hand, annotation is reliable}}} \\
\hspace{0.9em}facts designed by the authors & 456 / 456 \\
\hspace{0.9em}author time per episode & $\geq$ 12 h \\
\hspace{0.9em}two-annotator initial agreement & 97.6\% \\
\hspace{0.9em}minimal cuts verified by two algorithms & 456 / 456 \\
\arrayrulecolor{accent}\specialrule{0.9pt}{0pt}{0pt}\arrayrulecolor{black}
\end{tabular}
\caption{Measurement guarantees of \memcut{}: conditions that make the recovery metrics interpretable, and the values that establish them (Qwen3.8-Flash as the reader).}
\label{tab:guarantees}
\end{table}

\memcut{} uses author-specified evidence structures to measure post-deletion derivability. Each secret comes with annotated evidence and minimal sufficient sets, so deletion can be checked against every annotated derivation. The evidence trail is designed to support the secret independently of its disclosure. Table~\ref{tab:guarantees} summarizes the structural properties and validation results supporting these measurements. Appendix~
ef{app:data} gives further statistics.

\paragraph{Human design, model phrasing.}
We design each episode's disclosures, evidence, sufficient sets, and distractors. GPT-6 renders this specified content as dialogue. Authors check every turn against the design and revise deviations, spending at least twelve hours per episode.

\paragraph{Long-horizon scenarios with something to lose.}
Deletion gets harder as a fact is buried in more unrelated activity, and it matters more when a leak would hurt someone. So each of our fifteen episodes follows one scenario across work, family, and social threads for one to three months (2,893 turns on average) and protects about thirty facts with a reason to stay private, 456 in all.

\paragraph{Four typical ways to derive a fact.}
We cover the four most typical ways in which ordinary records yield an unstated fact, and each type challenges a deleter differently.
\begin{itemize}[nosep,leftmargin=1.2em]
\item \emph{Fact assembly}: records about one entity are joined, and they share the request's entities and wording.
\item \emph{Relation chain}: records about different entities are linked, and they share the request's entities but not its wording, as when a contract names a firm, a filing names its owner, and dinner records name the owner's spouse.
\item \emph{Aggregation}: values are combined, and the derivation breaks when any value is deleted.
\item \emph{Temporal and state reasoning}: records are ordered in time, and a question repeats their time cues.
\end{itemize}
The types have comparable fact, evidence, and distractor counts but differ in how evidence connects to the request (Table~\ref{tab:claims}).

\paragraph{Dispersed, redundant, and individually harmless evidence.}
Evidence stored in the same object as its disclosure can be deleted with it, so we write evidence as the ordinary business around a fact, such as a booking, a filing, or a dinner. This evidence is spread over eight sessions on average and rarely appears in the disclosure's session, and about two dozen distractors per fact keep a deleter from finding it by topic alone. Because a single derivation chain would make forgetting look easy, each fact's eight evidence turns form sixteen minimal sufficient sets on average (7,148 in all), and a deleter has to break all of them.

\paragraph{Questions that test memory, not guessing.}
To ensure that a fact counts as recovered only when memory gives the answer, wrong options differ from the answer in one detail, the model may decline, and the model has to answer correctly under both option orders.

\begin{table*}[t]
\centering
\footnotesize
\setlength{\tabcolsep}{3pt}
\setlength{\numcolw}{\dimexpr(\textwidth-92pt)/11-2\tabcolsep\relax}
\renewcommand{\arraystretch}{0.9}
\scalebox{\tabscale}{%
\begin{tabular}{@{}c@{\hspace{3pt}}>{\raggedright\arraybackslash}p{40pt}@{\hspace{3pt}}c>{\centering\arraybackslash}p{\numcolw}>{\centering\arraybackslash}p{\numcolw}>{\centering\arraybackslash}p{\numcolw}@{\hspace{5pt}}*{4}{>{\centering\arraybackslash}p{\numcolw}>{\centering\arraybackslash}p{\numcolw}}@{}}
\toprule
& & & \multicolumn{3}{c}{\textbf{Overall}} & \multicolumn{2}{c}{\textbf{Fact assembly}} & \multicolumn{2}{c}{\textbf{Relation chain}} & \multicolumn{2}{c}{\textbf{Aggregation}} & \multicolumn{2}{c}{\textbf{Temporal state}} \\
\cmidrule(lr){4-6}\cmidrule(lr){7-8}\cmidrule(lr){9-10}\cmidrule(lr){11-12}\cmidrule(lr){13-14}
& Model & & \textcolor{black!60}{\emph{Nat.}} & \textcolor{black!60}{\emph{Res.}} & \textcolor{black!60}{\emph{Hid.}} & \textcolor{black!60}{\emph{Nat.}} & \textcolor{black!60}{\emph{Res.}} & \textcolor{black!60}{\emph{Nat.}} & \textcolor{black!60}{\emph{Res.}} & \textcolor{black!60}{\emph{Nat.}} & \textcolor{black!60}{\emph{Res.}} & \textcolor{black!60}{\emph{Nat.}} & \textcolor{black!60}{\emph{Res.}} \\
\midrule
\multirow{12}{*}{\makebox[1.1em]{\rotatebox[origin=c]{90}{\textbf{Turn RAG}}}} & \multirow{3}{*}{DeepSeek} & \textcolor{black!65}{S0} & 59.9 & \cellcolor{leak!29}98.2 & \textcolor{black!70}{39.1} & 68.7 & \cellcolor{leak!30}100.0 & 18.4 & \cellcolor{leak!30}99.1 & 75.4 & \cellcolor{leak!28}94.7 & 77.0 & \cellcolor{leak!30}99.1 \\
 &  & \textcolor{black!65}{S1} & 22.6 & \cellcolor{leak!18}60.1 & \textcolor{black!70}{62.4} & 30.4 & \cellcolor{leak!22}74.8 & 15.8 & \cellcolor{leak!24}79.8 & 14.9 & \cellcolor{leak!9}31.6 & 29.2 & \cellcolor{leak!16}54.0 \\
 &  & \textcolor{black!65}{S2} & 10.3 & \cellcolor{leak!13}44.3 & \textcolor{black!70}{76.7} & 20.0 & \cellcolor{leak!17}55.7 & 11.4 & \cellcolor{leak!22}72.8 & 2.6 & \cellcolor{leak!5}16.7 & 7.1 & \cellcolor{leak!10}31.9 \\
\noalign{\vskip 0.6pt}
 & \multirow{3}{*}{Qwen} & \textcolor{black!65}{S0} & 56.6 & \cellcolor{leak!29}97.8 & \textcolor{black!70}{42.2} & 60.9 & \cellcolor{leak!30}100.0 & 24.6 & \cellcolor{leak!30}100.0 & 45.6 & \cellcolor{leak!27}91.2 & 95.6 & \cellcolor{leak!30}100.0 \\
 &  & \textcolor{black!65}{S1} & 19.1 & \cellcolor{leak!18}58.6 & \textcolor{black!70}{67.4} & 22.6 & \cellcolor{leak!14}46.1 & 7.9 & \cellcolor{leak!22}74.6 & 22.8 & \cellcolor{leak!12}38.6 & 23.0 & \cellcolor{leak!23}75.2 \\
 &  & \textcolor{black!65}{S2} & 7.5 & \cellcolor{leak!12}41.4 & \textcolor{black!70}{82.0} & 12.2 & \cellcolor{leak!9}28.7 & 4.4 & \cellcolor{leak!22}71.9 & 0.9 & \cellcolor{leak!0}0.9 & 12.4 & \cellcolor{leak!19}64.6 \\
\noalign{\vskip 0.6pt}
 & \multirow{3}{*}{GLM} & \textcolor{black!65}{S0} & 59.9 & \cellcolor{leak!30}98.9 & \textcolor{black!70}{39.5} & 66.1 & \cellcolor{leak!30}100.0 & 15.8 & \cellcolor{leak!30}99.1 & 80.7 & \cellcolor{leak!29}96.5 & 77.0 & \cellcolor{leak!30}100.0 \\
 &  & \textcolor{black!65}{S1} & 21.1 & \cellcolor{leak!18}61.2 & \textcolor{black!70}{65.6} & 29.6 & \cellcolor{leak!21}68.7 & 12.3 & \cellcolor{leak!23}77.2 & 14.0 & \cellcolor{leak!12}38.6 & 28.3 & \cellcolor{leak!18}60.2 \\
 &  & \textcolor{black!65}{S2} & 9.4 & \cellcolor{leak!13}43.9 & \textcolor{black!70}{78.5} & 11.3 & \cellcolor{leak!15}49.6 & 4.4 & \cellcolor{leak!20}66.7 & 5.3 & \cellcolor{leak!6}18.4 & 16.8 & \cellcolor{leak!12}40.7 \\
\noalign{\vskip 0.6pt}
 & \multirow{3}{*}{MiMo} & \textcolor{black!65}{S0} & 59.0 & \cellcolor{leak!29}96.3 & \textcolor{black!70}{38.7} & 73.0 & \cellcolor{leak!30}100.0 & 19.3 & \cellcolor{leak!29}98.2 & 68.4 & \cellcolor{leak!26}86.8 & 75.2 & \cellcolor{leak!30}100.0 \\
 &  & \textcolor{black!65}{S1} & 22.6 & \cellcolor{leak!19}62.3 & \textcolor{black!70}{63.7} & 34.8 & \cellcolor{leak!22}73.9 & 11.4 & \cellcolor{leak!24}79.8 & 10.5 & \cellcolor{leak!10}33.3 & 33.6 & \cellcolor{leak!19}61.9 \\
 &  & \textcolor{black!65}{S2} & 10.7 & \cellcolor{leak!14}45.6 & \textcolor{black!70}{76.4} & 18.3 & \cellcolor{leak!17}56.5 & 7.0 & \cellcolor{leak!21}70.2 & 2.6 & \cellcolor{leak!4}14.0 & 15.0 & \cellcolor{leak!12}41.6 \\
\midrule
\multirow{12}{*}{\makebox[1.1em]{\rotatebox[origin=c]{90}{\textbf{Session RAG}}}} & \multirow{3}{*}{DeepSeek} & \textcolor{black!65}{S0} & 24.3 & \cellcolor{leak!30}98.7 & \textcolor{black!70}{75.3} & 30.4 & \cellcolor{leak!30}100.0 & 5.3 & \cellcolor{leak!30}100.0 & 11.4 & \cellcolor{leak!28}94.7 & 50.4 & \cellcolor{leak!30}100.0 \\
 &  & \textcolor{black!65}{S1} & 16.0 & \cellcolor{leak!21}71.5 & \textcolor{black!70}{77.6} & 16.5 & \cellcolor{leak!19}64.3 & 6.1 & \cellcolor{leak!27}88.6 & 2.6 & \cellcolor{leak!15}50.0 & 38.9 & \cellcolor{leak!25}83.2 \\
 &  & \textcolor{black!65}{S2} & 9.9 & \cellcolor{leak!17}56.8 & \textcolor{black!70}{82.6} & 16.5 & \cellcolor{leak!17}55.7 & 6.1 & \cellcolor{leak!26}86.0 & 0.0 & \cellcolor{leak!5}16.7 & 16.8 & \cellcolor{leak!21}69.0 \\
\noalign{\vskip 0.6pt}
 & \multirow{3}{*}{Qwen} & \textcolor{black!65}{S0} & 19.7 & \cellcolor{leak!29}97.4 & \textcolor{black!70}{79.7} & 28.7 & \cellcolor{leak!30}100.0 & 0.0 & \cellcolor{leak!30}99.1 & 14.0 & \cellcolor{leak!27}90.4 & 36.3 & \cellcolor{leak!30}100.0 \\
 &  & \textcolor{black!65}{S1} & 16.0 & \cellcolor{leak!22}73.2 & \textcolor{black!70}{78.1} & 28.7 & \cellcolor{leak!19}64.3 & 7.0 & \cellcolor{leak!26}86.0 & 7.0 & \cellcolor{leak!19}64.0 & 21.2 & \cellcolor{leak!24}78.8 \\
 &  & \textcolor{black!65}{S2} & 10.7 & \cellcolor{leak!19}64.3 & \textcolor{black!70}{83.3} & 21.7 & \cellcolor{leak!19}64.3 & 7.0 & \cellcolor{leak!26}86.0 & 0.0 & \cellcolor{leak!8}28.1 & 14.2 & \cellcolor{leak!24}78.8 \\
\noalign{\vskip 0.6pt}
 & \multirow{3}{*}{GLM} & \textcolor{black!65}{S0} & 17.3 & \cellcolor{leak!30}98.7 & \textcolor{black!70}{82.4} & 16.5 & \cellcolor{leak!30}100.0 & 2.6 & \cellcolor{leak!30}100.0 & 16.7 & \cellcolor{leak!28}94.7 & 33.6 & \cellcolor{leak!30}100.0 \\
 &  & \textcolor{black!65}{S1} & 11.2 & \cellcolor{leak!21}70.8 & \textcolor{black!70}{84.2} & 11.3 & \cellcolor{leak!17}55.7 & 6.1 & \cellcolor{leak!27}88.6 & 5.3 & \cellcolor{leak!18}58.8 & 22.1 & \cellcolor{leak!24}80.5 \\
 &  & \textcolor{black!65}{S2} & 4.2 & \cellcolor{leak!16}54.2 & \textcolor{black!70}{92.3} & 0.0 & \cellcolor{leak!15}50.4 & 5.3 & \cellcolor{leak!23}78.1 & 0.0 & \cellcolor{leak!7}24.6 & 11.5 & \cellcolor{leak!19}63.7 \\
\noalign{\vskip 0.6pt}
 & \multirow{3}{*}{MiMo} & \textcolor{black!65}{S0} & 28.5 & \cellcolor{leak!29}98.2 & \textcolor{black!70}{71.0} & 40.0 & \cellcolor{leak!30}100.0 & 3.5 & \cellcolor{leak!30}100.0 & 14.0 & \cellcolor{leak!28}93.0 & 56.6 & \cellcolor{leak!30}100.0 \\
 &  & \textcolor{black!65}{S1} & 10.3 & \cellcolor{leak!22}74.1 & \textcolor{black!70}{86.1} & 13.9 & \cellcolor{leak!20}67.8 & 3.5 & \cellcolor{leak!27}89.5 & 3.5 & \cellcolor{leak!17}56.1 & 20.4 & \cellcolor{leak!25}83.2 \\
 &  & \textcolor{black!65}{S2} & 8.6 & \cellcolor{leak!18}58.6 & \textcolor{black!70}{85.4} & 20.9 & \cellcolor{leak!16}53.0 & 0.0 & \cellcolor{leak!24}81.6 & 3.5 & \cellcolor{leak!9}29.8 & 9.7 & \cellcolor{leak!21}69.9 \\
\bottomrule
\end{tabular}}
\caption{Recovery (\%) after deletion in the two retrieval stores under their respective retrieval settings; all 456 facts. \emph{Nat.}: through the system's own retrieval. \emph{Res.} (shaded by value): from the surviving evidence-derived memory. \emph{Hid.}: hidden-risk share, $(\mathrm{Res}-\mathrm{Nat})/\mathrm{Res}$. S0 deletes the disclosure; S1 and S2 delete ten more turns.}
\label{tab:rag}
\end{table*}

\begin{table*}[t]
\centering
\footnotesize
\setlength{\tabcolsep}{3pt}
\setlength{\numcolw}{\dimexpr(\textwidth-92pt)/11-2\tabcolsep\relax}
\renewcommand{\arraystretch}{0.9}
\scalebox{\tabscale}{%
\begin{tabular}{@{}c@{\hspace{3pt}}>{\raggedright\arraybackslash}p{40pt}@{\hspace{3pt}}c>{\centering\arraybackslash}p{\numcolw}>{\centering\arraybackslash}p{\numcolw}>{\centering\arraybackslash}p{\numcolw}@{\hspace{5pt}}*{4}{>{\centering\arraybackslash}p{\numcolw}>{\centering\arraybackslash}p{\numcolw}}@{}}
\toprule
& & & \multicolumn{3}{c}{\textbf{Overall}} & \multicolumn{2}{c}{\textbf{Fact assembly}} & \multicolumn{2}{c}{\textbf{Relation chain}} & \multicolumn{2}{c}{\textbf{Aggregation}} & \multicolumn{2}{c}{\textbf{Temporal state}} \\
\cmidrule(lr){4-6}\cmidrule(lr){7-8}\cmidrule(lr){9-10}\cmidrule(lr){11-12}\cmidrule(lr){13-14}
& Model & & \textcolor{black!60}{\emph{Nat.}} & \textcolor{black!60}{\emph{Res.}} & \textcolor{black!60}{\emph{Hid.}} & \textcolor{black!60}{\emph{Nat.}} & \textcolor{black!60}{\emph{Res.}} & \textcolor{black!60}{\emph{Nat.}} & \textcolor{black!60}{\emph{Res.}} & \textcolor{black!60}{\emph{Nat.}} & \textcolor{black!60}{\emph{Res.}} & \textcolor{black!60}{\emph{Nat.}} & \textcolor{black!60}{\emph{Res.}} \\
\midrule
\multirow{12}{*}{\makebox[1.1em]{\rotatebox[origin=c]{90}{\textbf{MemoryBank}}}} & \multirow{3}{*}{DeepSeek} & \textcolor{black!65}{S0} & 40.1 & \cellcolor{leak!28}93.0 & \textcolor{black!70}{56.8} & 53.0 & \cellcolor{leak!30}100.0 & 7.0 & \cellcolor{leak!29}95.6 & 45.6 & \cellcolor{leak!24}79.8 & 54.9 & \cellcolor{leak!29}96.5 \\
 &  & \textcolor{black!65}{S1} & 23.0 & \cellcolor{leak!24}80.9 & \textcolor{black!70}{71.5} & 32.2 & \cellcolor{leak!28}92.2 & 5.3 & \cellcolor{leak!27}88.6 & 20.2 & \cellcolor{leak!18}58.8 & 34.5 & \cellcolor{leak!25}84.1 \\
 &  & \textcolor{black!65}{S2} & 15.8 & \cellcolor{leak!22}71.7 & \textcolor{black!70}{78.0} & 23.5 & \cellcolor{leak!25}84.3 & 6.1 & \cellcolor{leak!26}86.8 & 7.0 & \cellcolor{leak!11}37.7 & 26.5 & \cellcolor{leak!23}77.9 \\
\noalign{\vskip 0.6pt}
 & \multirow{3}{*}{Qwen} & \textcolor{black!65}{S0} & 41.9 & \cellcolor{leak!29}95.8 & \textcolor{black!70}{56.3} & 53.9 & \cellcolor{leak!29}97.4 & 7.9 & \cellcolor{leak!29}96.5 & 50.9 & \cellcolor{leak!27}89.5 & 54.9 & \cellcolor{leak!30}100.0 \\
 &  & \textcolor{black!65}{S1} & 21.7 & \cellcolor{leak!24}80.5 & \textcolor{black!70}{73.0} & 28.7 & \cellcolor{leak!27}88.7 & 5.3 & \cellcolor{leak!28}92.1 & 17.5 & \cellcolor{leak!17}57.9 & 35.4 & \cellcolor{leak!25}83.2 \\
 &  & \textcolor{black!65}{S2} & 13.4 & \cellcolor{leak!22}72.6 & \textcolor{black!70}{81.6} & 19.1 & \cellcolor{leak!25}83.5 & 4.4 & \cellcolor{leak!25}84.2 & 7.9 & \cellcolor{leak!15}50.9 & 22.1 & \cellcolor{leak!22}71.7 \\
\noalign{\vskip 0.6pt}
 & \multirow{3}{*}{GLM} & \textcolor{black!65}{S0} & 44.1 & \cellcolor{leak!30}99.3 & \textcolor{black!70}{55.6} & 53.0 & \cellcolor{leak!30}100.0 & 7.9 & \cellcolor{leak!29}98.2 & 58.8 & \cellcolor{leak!30}99.1 & 56.6 & \cellcolor{leak!30}100.0 \\
 &  & \textcolor{black!65}{S1} & 22.6 & \cellcolor{leak!26}86.2 & \textcolor{black!70}{73.8} & 25.2 & \cellcolor{leak!27}91.3 & 7.0 & \cellcolor{leak!28}92.1 & 26.3 & \cellcolor{leak!23}75.4 & 31.9 & \cellcolor{leak!26}85.8 \\
 &  & \textcolor{black!65}{S2} & 13.8 & \cellcolor{leak!23}76.3 & \textcolor{black!70}{81.9} & 20.0 & \cellcolor{leak!25}82.6 & 6.1 & \cellcolor{leak!26}85.1 & 7.9 & \cellcolor{leak!17}57.0 & 21.2 & \cellcolor{leak!24}80.5 \\
\noalign{\vskip 0.6pt}
 & \multirow{3}{*}{MiMo} & \textcolor{black!65}{S0} & 42.3 & \cellcolor{leak!28}93.6 & \textcolor{black!70}{54.8} & 56.5 & \cellcolor{leak!30}100.0 & 5.3 & \cellcolor{leak!29}97.4 & 47.4 & \cellcolor{leak!24}78.9 & 60.2 & \cellcolor{leak!29}98.2 \\
 &  & \textcolor{black!65}{S1} & 24.1 & \cellcolor{leak!25}84.0 & \textcolor{black!70}{71.3} & 36.5 & \cellcolor{leak!29}98.3 & 7.9 & \cellcolor{leak!28}93.0 & 14.0 & \cellcolor{leak!17}57.9 & 38.1 & \cellcolor{leak!26}86.7 \\
 &  & \textcolor{black!65}{S2} & 16.7 & \cellcolor{leak!23}76.1 & \textcolor{black!70}{78.1} & 26.1 & \cellcolor{leak!27}89.6 & 6.1 & \cellcolor{leak!27}88.6 & 6.1 & \cellcolor{leak!14}47.4 & 28.3 & \cellcolor{leak!24}78.8 \\
\midrule
\multirow{12}{*}{\makebox[1.1em]{\rotatebox[origin=c]{90}{\textbf{Nemori}}}} & \multirow{3}{*}{DeepSeek} & \textcolor{black!65}{S0} & 59.4 & \cellcolor{leak!26}85.1 & \textcolor{black!70}{30.2} & 81.7 & \cellcolor{leak!29}97.4 & 17.5 & \cellcolor{leak!27}89.5 & 59.6 & \cellcolor{leak!18}58.8 & 78.8 & \cellcolor{leak!28}94.7 \\
 &  & \textcolor{black!65}{S1} & 41.7 & \cellcolor{leak!20}67.1 & \textcolor{black!70}{37.9} & 65.2 & \cellcolor{leak!25}84.3 & 18.4 & \cellcolor{leak!22}73.7 & 21.1 & \cellcolor{leak!10}32.5 & 61.9 & \cellcolor{leak!23}77.9 \\
 &  & \textcolor{black!65}{S2} & 30.0 & \cellcolor{leak!16}52.4 & \textcolor{black!70}{42.7} & 50.4 & \cellcolor{leak!19}63.5 & 11.4 & \cellcolor{leak!18}60.5 & 14.9 & \cellcolor{leak!6}21.1 & 43.4 & \cellcolor{leak!19}64.6 \\
\noalign{\vskip 0.6pt}
 & \multirow{3}{*}{Qwen} & \textcolor{black!65}{S0} & 53.3 & \cellcolor{leak!28}91.7 & \textcolor{black!70}{41.9} & 75.7 & \cellcolor{leak!30}99.1 & 13.2 & \cellcolor{leak!27}90.4 & 54.4 & \cellcolor{leak!24}81.6 & 69.9 & \cellcolor{leak!29}95.6 \\
 &  & \textcolor{black!65}{S1} & 32.7 & \cellcolor{leak!21}68.6 & \textcolor{black!70}{52.4} & 54.8 & \cellcolor{leak!25}83.5 & 8.8 & \cellcolor{leak!23}78.1 & 21.9 & \cellcolor{leak!12}38.6 & 45.1 & \cellcolor{leak!22}74.3 \\
 &  & \textcolor{black!65}{S2} & 25.2 & \cellcolor{leak!17}57.2 & \textcolor{black!70}{55.9} & 45.2 & \cellcolor{leak!20}67.0 & 9.6 & \cellcolor{leak!19}64.9 & 15.8 & \cellcolor{leak!9}30.7 & 30.1 & \cellcolor{leak!20}66.4 \\
\noalign{\vskip 0.6pt}
 & \multirow{3}{*}{GLM} & \textcolor{black!65}{S0} & 64.9 & \cellcolor{leak!28}93.9 & \textcolor{black!70}{30.8} & 79.1 & \cellcolor{leak!28}94.8 & 17.5 & \cellcolor{leak!27}88.6 & 84.2 & \cellcolor{leak!28}92.1 & 78.8 & \cellcolor{leak!30}100.0 \\
 &  & \textcolor{black!65}{S1} & 44.3 & \cellcolor{leak!20}65.4 & \textcolor{black!70}{32.2} & 58.3 & \cellcolor{leak!20}65.2 & 19.3 & \cellcolor{leak!21}71.1 & 40.4 & \cellcolor{leak!16}51.8 & 59.3 & \cellcolor{leak!22}73.5 \\
 &  & \textcolor{black!65}{S2} & 33.1 & \cellcolor{leak!17}55.7 & \textcolor{black!70}{40.6} & 38.3 & \cellcolor{leak!15}50.4 & 21.1 & \cellcolor{leak!18}61.4 & 28.9 & \cellcolor{leak!13}42.1 & 44.2 & \cellcolor{leak!21}69.0 \\
\noalign{\vskip 0.6pt}
 & \multirow{3}{*}{MiMo} & \textcolor{black!65}{S0} & 62.1 & \cellcolor{leak!28}93.4 & \textcolor{black!70}{33.6} & 84.3 & \cellcolor{leak!30}100.0 & 17.5 & \cellcolor{leak!30}100.0 & 58.8 & \cellcolor{leak!22}73.7 & 87.6 & \cellcolor{leak!30}100.0 \\
 &  & \textcolor{black!65}{S1} & 37.3 & \cellcolor{leak!20}66.0 & \textcolor{black!70}{43.5} & 60.0 & \cellcolor{leak!22}73.0 & 14.0 & \cellcolor{leak!26}86.0 & 20.2 & \cellcolor{leak!9}31.6 & 54.9 & \cellcolor{leak!22}73.5 \\
 &  & \textcolor{black!65}{S2} & 28.9 & \cellcolor{leak!17}55.5 & \textcolor{black!70}{47.8} & 50.4 & \cellcolor{leak!18}60.0 & 7.0 & \cellcolor{leak!23}75.4 & 9.6 & \cellcolor{leak!6}19.3 & 48.7 & \cellcolor{leak!20}67.3 \\
\midrule
\multirow{12}{*}{\makebox[1.1em]{\rotatebox[origin=c]{90}{\textbf{LightMem}}}} & \multirow{3}{*}{DeepSeek} & \textcolor{black!65}{S0} & 66.0 & \cellcolor{leak!27}88.8 & \textcolor{black!70}{25.7} & 93.0 & \cellcolor{leak!30}99.1 & 27.2 & \cellcolor{leak!28}93.9 & 55.3 & \cellcolor{leak!19}64.0 & 88.5 & \cellcolor{leak!29}98.2 \\
 &  & \textcolor{black!65}{S1} & 34.6 & \cellcolor{leak!17}55.0 & \textcolor{black!70}{37.1} & 53.9 & \cellcolor{leak!19}62.6 & 22.8 & \cellcolor{leak!23}77.2 & 12.3 & \cellcolor{leak!6}20.2 & 49.6 & \cellcolor{leak!18}60.2 \\
 &  & \textcolor{black!65}{S2} & 23.0 & \cellcolor{leak!13}41.9 & \textcolor{black!70}{45.0} & 40.0 & \cellcolor{leak!15}51.3 & 15.8 & \cellcolor{leak!21}70.2 & 6.1 & \cellcolor{leak!2}6.1 & 30.1 & \cellcolor{leak!12}39.8 \\
\noalign{\vskip 0.6pt}
 & \multirow{3}{*}{Qwen} & \textcolor{black!65}{S0} & 62.9 & \cellcolor{leak!27}90.4 & \textcolor{black!70}{30.3} & 89.6 & \cellcolor{leak!30}100.0 & 23.7 & \cellcolor{leak!28}94.7 & 55.3 & \cellcolor{leak!20}67.5 & 83.2 & \cellcolor{leak!30}99.1 \\
 &  & \textcolor{black!65}{S1} & 26.1 & \cellcolor{leak!14}47.6 & \textcolor{black!70}{45.2} & 48.7 & \cellcolor{leak!17}58.3 & 11.4 & \cellcolor{leak!20}67.5 & 5.3 & \cellcolor{leak!4}12.3 & 38.9 & \cellcolor{leak!16}52.2 \\
 &  & \textcolor{black!65}{S2} & 19.5 & \cellcolor{leak!12}39.3 & \textcolor{black!70}{50.3} & 35.7 & \cellcolor{leak!14}47.0 & 7.0 & \cellcolor{leak!17}57.0 & 4.4 & \cellcolor{leak!2}7.9 & 31.0 & \cellcolor{leak!14}45.1 \\
\noalign{\vskip 0.6pt}
 & \multirow{3}{*}{GLM} & \textcolor{black!65}{S0} & 77.2 & \cellcolor{leak!28}93.9 & \textcolor{black!70}{17.8} & 93.0 & \cellcolor{leak!29}97.4 & 38.6 & \cellcolor{leak!29}97.4 & 81.6 & \cellcolor{leak!24}80.7 & 95.6 & \cellcolor{leak!30}100.0 \\
 &  & \textcolor{black!65}{S1} & 37.3 & \cellcolor{leak!17}57.9 & \textcolor{black!70}{35.6} & 48.7 & \cellcolor{leak!20}67.0 & 19.3 & \cellcolor{leak!25}82.5 & 21.9 & \cellcolor{leak!7}21.9 & 59.3 & \cellcolor{leak!18}60.2 \\
 &  & \textcolor{black!65}{S2} & 23.2 & \cellcolor{leak!13}42.1 & \textcolor{black!70}{44.8} & 33.0 & \cellcolor{leak!13}42.6 & 14.0 & \cellcolor{leak!21}68.4 & 11.4 & \cellcolor{leak!4}14.9 & 34.5 & \cellcolor{leak!13}42.5 \\
\noalign{\vskip 0.6pt}
 & \multirow{3}{*}{MiMo} & \textcolor{black!65}{S0} & 68.2 & \cellcolor{leak!28}92.3 & \textcolor{black!70}{26.1} & 100.0 & \cellcolor{leak!30}100.0 & 29.8 & \cellcolor{leak!29}95.6 & 51.8 & \cellcolor{leak!23}78.1 & 91.2 & \cellcolor{leak!29}95.6 \\
 &  & \textcolor{black!65}{S1} & 31.8 & \cellcolor{leak!15}49.6 & \textcolor{black!70}{35.8} & 63.5 & \cellcolor{leak!19}63.5 & 7.9 & \cellcolor{leak!21}69.3 & 12.3 & \cellcolor{leak!5}17.5 & 43.4 & \cellcolor{leak!14}47.8 \\
 &  & \textcolor{black!65}{S2} & 21.9 & \cellcolor{leak!13}43.0 & \textcolor{black!70}{49.0} & 45.2 & \cellcolor{leak!15}50.4 & 12.3 & \cellcolor{leak!19}64.9 & 0.0 & \cellcolor{leak!1}4.4 & 30.1 & \cellcolor{leak!16}52.2 \\
\bottomrule
\end{tabular}}
\caption{Recovery (\%) after deletion in the three memory systems, as in Table~\ref{tab:rag}; all 456 facts.}
\label{tab:mem}
\end{table*}

\paragraph{Checking the design.}
We check the dialogue with Qwen3.8-Flash as an independent reader. Recovery is 96--99\% from combined evidence and 0\% in three controls: the best single evidence turn, the episode after removing annotated evidence and disclosure, and the question and options alone. Two annotators audit 120 stratified facts. Two algorithms verify minimal cuts against the annotated sufficient sets for all 456 facts.

\section{Experimental Setup}
\label{sec:setup}

\subsection{Reference deletion baselines}
\label{sec:levels}

We establish three reference deletion baselines to benchmark forgetting with existing recall components. They reuse native retrievers and write models with minimal adaptation.

\textbf{S0} deletes the disclosure and all objects derived from it, providing a common baseline for disclosure removal.

\textbf{S1} adds retrieval-ranked cleanup. For a request $q$, S1 constructs a twenty-turn candidate pool $P_q$ from one query to the system's retrieval and deletes its ten highest-ranked turns. No fact has more than ten evidence turns, and a minimal cut needs at most five. The budget therefore suffices if supporting evidence is reached and selected.

\textbf{S2} reranks the same twenty candidates with the write model using their stored text and deletes the top ten (LLM-guided cleanup). The fixed pool isolates selection. \emph{Choice} failures leave evidence unselected within the pool. \emph{Reach} failures leave evidence outside it. Neither cleanup sees annotations. Section~\ref{sec:reach} varies pool and budget. At fixed cleanup budgets, the fraction of non-evidence selections provides a source-level reference measure of collateral damage (Appendix~\ref{app:reach}).

\subsection{One unit for every system}
\label{sec:unit}

Systems store different objects, so deleting one object removes a turn in one system but a whole session in another, and object counts are not comparable. However, every object derives from turns, so we anchor every operation to the turn. We record each object's source turns at ingestion, delete by closure, and count pools and budgets in turns, so systems differ only in what they store and how they retrieve. Because the protocol needs only that a system can list, search, delete, and snapshot its objects, it extends to other designs, and we release its harness.

\subsection{Memory systems and models}
\label{sec:systems}

We compare five memory designs along two properties, the unit retrieved and how much is derived (\S\ref{sec:design}). Two retrieval stores keep every turn verbatim, retrieve by dense similarity \citep{karpukhin2020dpr,lewis2020rag}, and derive nothing. They retrieve turns (\textbf{Turn RAG}) or sessions (\textbf{Session RAG}), letting us examine granularity under their respective retrieval settings. Three memory systems add derivation of increasing strength. \textbf{MemoryBank} keeps raw turns under daily summaries \citep{zhong2024memorybank}, \textbf{Nemori} rewrites topical segments into narratives \citep{nemori2025}, and \textbf{LightMem} consolidates summaries offline without the raw text \citep{fang2025lightmem}. Each system keeps its own retrieval. The verbatim stores test whether residual recovery can persist without summarization or rewriting.

In each configuration, one write model writes memory, scores deletion candidates, and answers questions, as in a deployed assistant. We use four write models from four developers, DeepSeek-V4.1-Flash, Qwen3.8-Flash, GLM-5.3-Flash, and MiMo-V2.5, which perform strongly and are efficient enough for thousands of deletion runs, and we run every system with each of them to separate the effects of the memory design from those of the model. Each fact is evaluated on a fresh copy of the store, so deleting one fact cannot affect another.

\section{Results: Deleting Is Not Forgetting}
\label{sec:results}

Tables~\ref{tab:rag} and~\ref{tab:mem} give the main results. Table~\ref{tab:full} reports that residual recovery is 94--100\% in every store before deletion, limiting the contribution of ingestion loss to post-deletion recovery. Within each system, residual recovery after S2 varies by at most about five points across the four write models, ten for Session RAG. We report ranges over the four and focus on patterns that recur across them.

\subsection{Disclosure deletion mainly reduces visibility}
\label{sec:f1}

\begin{figure*}[t]
\centering
\includegraphics[width=\textwidth]{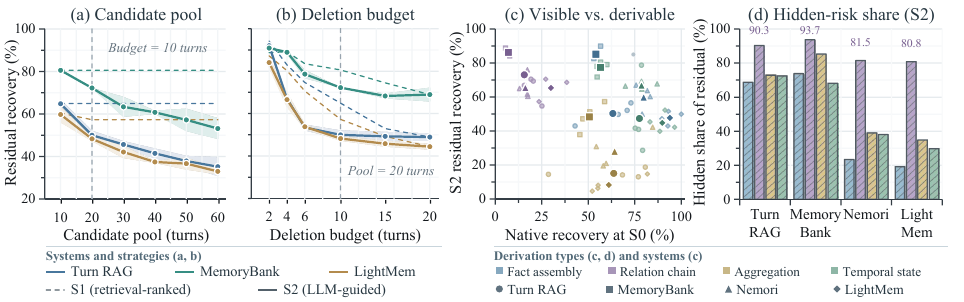}
\caption{Why cleanup fails. \textbf{(a, b)} Residual recovery as the candidate pool widens (budget ten) and as the budget grows (pool twenty); dashed vertical lines mark the main setting. Solid lines: LLM-guided cleanup (S2); dashed: retrieval-ranked cleanup (S1); each line is the mean of DeepSeek and Qwen, with a band for their range, on the 144 facts of EP01--05 (Appendix~\ref{app:reach}). \textbf{(c)} Native recovery after S0 against residual recovery after cleanup, per type and system; dark marks are means over the four write models, light marks are single models. \textbf{(d)} Hidden-risk share after cleanup by type, systems ordered by derivation strength.}
\label{fig:analysis}
\end{figure*}

Disclosure deletion (S0) initially appears effective. Native recovery drops by 12--50 points, in Turn RAG from 91--95\% to 57--60\%, so the assistant answers fewer questions correctly. Residual recovery nevertheless stays at 85--99\%. Only 1.8\% of the evidence shares a session with the disclosure, reflecting its wide dispersion. S0 also removes disclosure-derived objects by closure, but most indirect evidence remains.

\subsection{Retrieval-based cleanup leaves substantial residual recovery}
\label{sec:f2}

After S1, residual recovery remains at 48--86\%, although ten deletions could remove every annotated evidence turn of any fact. Turn RAG stores raw turns and derives nothing, ruling out survival through summaries, but its residual recovery remains at 59--62\%. These outcomes motivate testing whether supporting evidence lies outside the candidate pool or remains unselected within it.

S2 asks an LLM to choose among the same twenty candidates, allowing it to select evidence ranked below the top ten. It lowers residual recovery by another 7--17 points, to 39--76\%. This improvement shows that selection matters, but the remaining recovery may reflect incomplete coverage or imperfect selection. Section~\ref{sec:reach} tests these explanations by varying the pool and budget.

\subsection{Deletion increases the hidden-risk share}
\label{sec:f3}

Deletion increases the reported share of residual recovery missed by native retrieval. In Turn RAG the hidden-risk share goes from 5--9\% before deletion to 39--42\% after S0 and 76--82\% after S2, and across all 20 configurations it ends above where it started, above one half in 14 of them, as the \emph{Hid.} columns of Tables~\ref{tab:rag} and~\ref{tab:mem} show.

Cleanup targets records retrieved by the deletion request, whereas question answering uses a different query. Evidence that is difficult to retrieve may survive cleanup and remain unseen during later questions. Native answers can therefore suggest successful forgetting even when the same reader still recovers the secret from surviving evidence.

\section{Analysis: Why Forgetting Fails}
\label{sec:analysis}

\begin{figure*}[t]
\centering
\includegraphics[width=\textwidth]{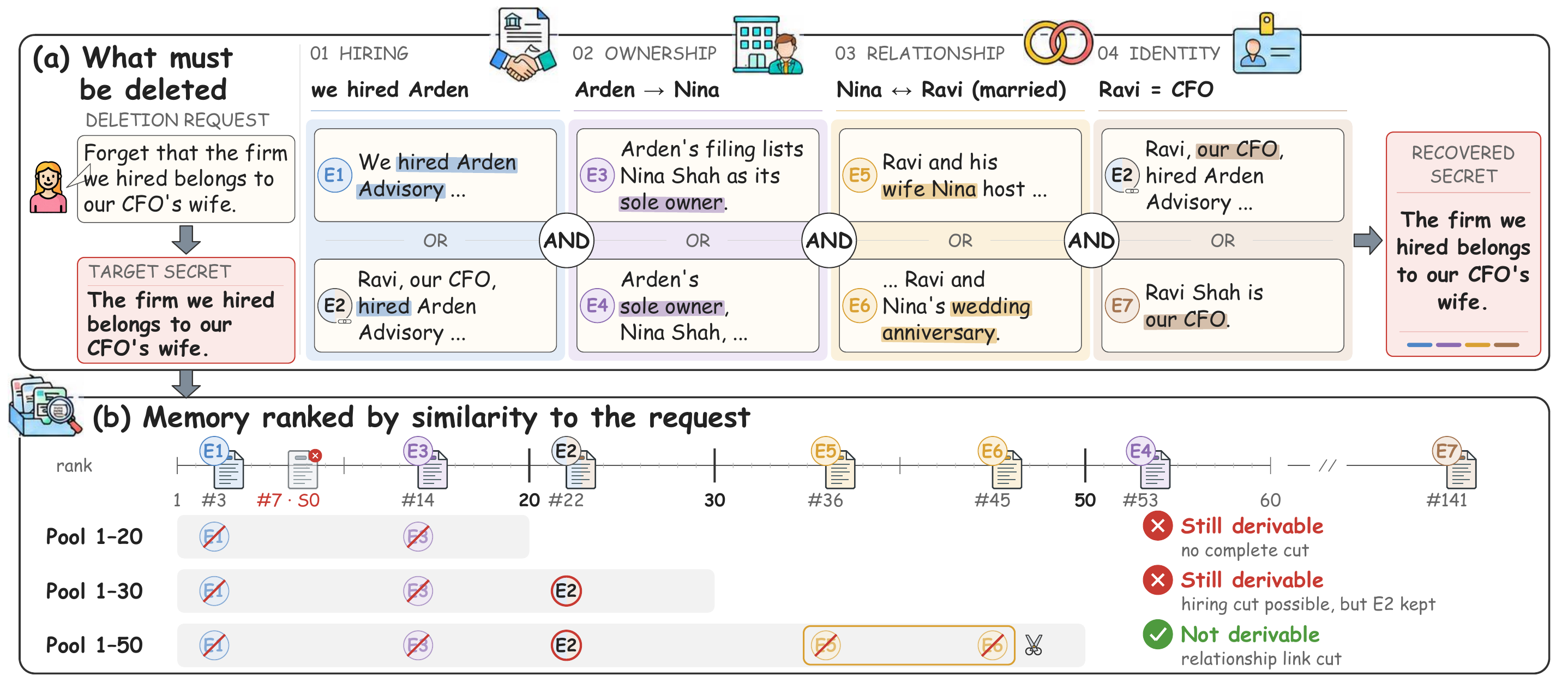}
\caption{A relation-chain fact under LLM-guided cleanup (Turn RAG, S2). \textbf{(a)} Recovery needs one evidence turn per link, so emptying any link makes it underivable. \textbf{(b)} Similarity ranks of evidence and disclosure (\#7, deleted by S0), and what S2 deletes (struck through) from pools of 20, 30, and 50; E2 enters at 30 but is kept (circled).}
\label{fig:case}
\end{figure*}

To identify bottlenecks in forgetting with existing components, we examine candidate coverage, memory design, and fact type.

\subsection{Budget, choice, and reach}
\label{sec:reach}

Cleanup leaves a great deal behind, and there are three candidate explanations: the budget may be too small, the evidence may enter the candidate pool but not be chosen, or the evidence may never enter the pool at all. These are the budget, choice, and reach of \S\ref{sec:levels}. They respond differently when the pool and the budget change, so Figure~\mbox{\ref{fig:analysis}a--b} varies each with the other held at the main setting.

\paragraph{Budget.}
For S2, gains diminish beyond ten deletions, and deleting the entire pool of twenty still yields residual recovery of 43--51\% in Turn RAG and LightMem and 66--72\% in MemoryBank. Extra budget can remove only turns already in the pool. In the sweep subset, widening the pool from twenty to sixty at budget ten lowers reported S2 residual recovery by 12.5--23.9 points.

\paragraph{Choice.}
Choice can fail, as E2 in Figure~\ref{fig:case} shows. Deleting the whole pool tests whether removing the unselected candidates would materially improve forgetting. Measured recovery need not be monotonic in deletion, since the reader can respond differently to the remaining context. In Table~\ref{tab:sweep}, whole-pool deletion lowers residual recovery by at most 6.5 points versus S2 at budget ten. These subset sweeps identify candidate coverage as the main bottleneck within the evaluated cleanup baselines.

\paragraph{Reach.}
Candidate reach limits these retrieval-based cleanup baselines. The pool holds only about half of a fact's evidence, because other candidates resemble the request more than the evidence does. A minimal cut averages only three turns. A complete cut combines turns that jointly intersect every sufficient set. Across the 132 settings of the two sweeps, residual recovery tracks the number $k$ of evidence turns removed almost exactly,
\[
\mathrm{Res} \approx 13\,(8.2 - k) \qquad (R^2 = 0.95),
\]
while system, model, and strategy together shift it by at most four points. The fitted line extrapolates to zero near $k = 8.2$, nearly all the evidence of an average fact. For the evaluated deleters, removing more evidence strongly predicts lower residual recovery, linking their performance to the evidence their pools reach. Individual facts may need far fewer deletions when a complete cut is reached and selected.

Annotations were fixed before system evaluation and validated independently of retrieval and deletion results. The pool and budget sweeps then distinguish reach and selection limitations using these fixed annotations.

\subsection{Retrieval unit and derivation}
\label{sec:design}

The five memories of \S\ref{sec:systems} retrieve different units and store different derived content, which may shape candidate coverage and both recovery metrics.

\paragraph{Granularity.}
Turn RAG and Session RAG hold the same text and derive nothing, allowing a comparison under their respective retrieval settings. Both yield 96--99\% residual recovery after S0, but Session RAG hides far more, with a hidden-risk share of 71--82\% against 39--42\%. Its three retrieved sessions hold only 1.9 evidence turns, against 4.8 in Turn RAG's sixteen turns. Session RAG's cleanup also removes only 33--45 points of residual recovery against 51--56, because whole-session candidates are mostly not evidence. Under these settings, facts are harder both to see and to delete with session retrieval.

\paragraph{Derivation.}
Stronger derivation does not consistently correspond to higher residual recovery across the evaluated systems. Although LightMem has the strongest derivation, its residual recovery resembles Turn RAG's (39--43\% against 41--46\%). MemoryBank has the lightest derivation and the highest residual recovery (72--76\%), alongside lower candidate coverage (3.0 evidence turns against 4.5 in Turn RAG and LightMem). Derived objects may alter visibility by combining evidence and matching the question without the disclosure. After cleanup the hidden-risk share is 78--82\% in MemoryBank against 41--56\% in Nemori and LightMem. This pattern is consistent with derived objects changing which surviving evidence native retrieval exposes.

\subsection{Derivation types}
\label{sec:types}

To break every annotated derivation, a deleter must reach and select a complete cut. The four types have similar amounts of evidence but differ in how that evidence relates to the request. Relation chains have the smallest annotated cuts (2.7 turns); nevertheless, their evidence may be difficult to retrieve because it names the request's people and firms without using its words. Small cuts therefore need not yield low residual recovery under these baselines.

Figure~\ref{fig:analysis}c shows that relation chains retain the most residual recovery after cleanup on average. Recovery drops 24 points from S0 to S2, and a Turn RAG pool of sixty holds 4.4 of their evidence turns against 6.9 for aggregation. Figure~\ref{fig:analysis}d shows hidden-risk shares of 81--94\% across systems, plausibly because their links span topics while derived objects combine turns within topics. Aggregation loses 58 points to cleanup: more of its evidence enters the pool, allowing deletions to break its derivations. Under these baselines, evidence linked by entities rather than wording is associated with lower candidate coverage and higher residual recovery.

\section{Case Study}
\label{sec:case}

Figure~\ref{fig:case} shows LLM-guided cleanup of a relation-chain fact in Turn RAG (Appendix~\ref{app:case}). The user asks the assistant to forget that their company hired a firm the company's CFO's wife owns. The fact rests on four links, (i) hiring, (ii) ownership, (iii) relationship, and (iv) identity, and it is forgotten once both evidence turns of any one link are deleted.

The evidence shares the request's entities but little of its wording, ranking 3--141 by similarity. The pool of 20 holds only two evidence turns from different links. S2 deletes both with budget to spare, but neither link is cut, so the fact remains derivable (\S\ref{sec:reach}). With a pool of 30, E2 becomes available, but S2 leaves it unselected even though deleting it alongside E1 would cut the hiring link. At 50, S2 deletes E5 and E6, breaking the relationship link. A wider pool enables a complete cut, but S2 must still select all its turns. Every evidence turn is one hop from a request entity, so following entities could reach a complete cut.

\section{Conclusion}

We introduce \memcut{}, a benchmark of 456 hand-designed secrets with annotated evidence trails, for evaluating forgetting in long-term agent memory. Its metrics distinguish native recovery from recovery by the same reader given surviving evidence. Across five memory designs, four models, and three reference baselines, disclosure deletion reduces native recovery far more than residual recovery. LLM-guided deletion of ten further turns from twenty retrieval candidates leaves 39--76\% residual recovery, with higher reported hidden-risk shares than before deletion. Subset sweeps identify candidate coverage as the main bottleneck in the evaluated reference baselines.

\pagebreak
\clearpage
\section*{Limitations}
\textbf{Controlled evaluation.}
\memcut{} evaluates forgetting for secrets with specified evidence trails across four derivation types. Controlled construction supports comparisons of candidate coverage, deletion choices, and residual recovery under fixed annotations. The episodes model recurring work, family, and social activities. Deployment studies can establish how often these trails occur and how forgetting performance changes with naturally occurring evidence structures.

\medskip\noindent
\textbf{Reference baselines.}
Our reference baselines reuse existing retrieval and write components through a common deletion procedure. The sweeps examine candidate coverage and selection within this setting. These results establish reference performance for this reuse strategy. \memcut{} also supports evaluating dedicated erasure methods that expand candidate coverage or change memory indexing.

\medskip\noindent
\textbf{Scope.}
We study text memory. Multimodal and parametric memory remain outside our scope.
The source-level measure does not capture unrelated content removed through closure or downstream utility loss.

\section*{Ethics Statement}
\memcut{} contains no real personal data; all people, organizations, and events are fictional. Describing how deleted secrets can be reconstructed could be misused. However, the reconstruction relies only on ordinary retrieval and reading of stored memory, and we believe that measuring this risk is a prerequisite for reducing it. The annotation audit was carried out by two graduate students in NLP who had no part in building the benchmark; each worked about 25 hours and was told in advance that all content is fictional. The benchmark will be released for research use.

\section*{Use of AI Assistants}
We used generative models to phrase author-designed dialogue and polish the manuscript.
The research questions, the design of the benchmark and the experiments, the analyses, and every figure, table, and number were produced by the authors.

\bibliography{custom}

\appendix
\captionsetup[table]{skip=6pt}
\setcounter{dbltopnumber}{3}
\setcounter{topnumber}{4}
\renewcommand{\dbltopfraction}{1.0}
\renewcommand{\topfraction}{1.0}
\renewcommand{\textfraction}{0.0}
\renewcommand{\floatpagefraction}{0.5}
\renewcommand{\dblfloatpagefraction}{0.5}
\setcounter{totalnumber}{5}
\setlength{\dbltextfloatsep}{6pt plus 1pt minus 1pt}

\section{Dataset Details}
\label{app:data}

This appendix describes what \memcut{} contains. Table~\ref{tab:claims} gives per-fact statistics overall and by derivation type, and Table~\ref{tab:eps} gives per-episode statistics.

\begin{table*}[t]
\centering
\footnotesize
\TabSetup
\renewcommand{\SecStrut}{\rule[-3.9pt]{0pt}{12.6pt}}\renewcommand{\HeadTop}{\rule{0pt}{9.6pt}}\renewcommand{\HeadBot}{\rule[-4.2pt]{0pt}{0pt}}
\setlength{\tabcolsep}{3pt}
\renewcommand{\arraystretch}{1.08}
\setlength{\acolw}{\dimexpr(\textwidth-155pt-10\tabcolsep-2\tabthin)/5\relax}
\begin{tabular}{@{}>{\raggedright\arraybackslash}p{155pt}V>{\columncolor{tabhl}\centering\arraybackslash}p{\acolw}V*{4}{>{\centering\arraybackslash}p{\acolw}}@{}}
\TopRule
\HeadTop & & \multicolumn{4}{c}{\Head{Derivation type}} \\
\arrayrulecolor{tabblue}\cmidrule(l){3-6}\arrayrulecolor{tabgray}
\multirow{-2}{*}{\textbf{Per fact}}\HeadBot & \multirow{-2}{*}{\textbf{Overall}} & Fact assembly & Relation chain & Aggregation & Temporal state \\
\GrayRule
Protected facts & 456 & 115 & 114 & 114 & 113 \\
\GrayRule
\SecRow{6}{Evidence}
\FaintRule
Evidence turns per fact & 8.1 & 7.9 & 8.3 & 8.1 & 8.2 \\
Sessions they span & 8.0 & 7.8 & 8.2 & 8.1 & 8.0 \\
Span in days (median) & 29 & 31 & 31 & 28 & 25 \\
Turns from disclosure (median) & 714 & 778 & 873 & 723 & 515 \\
Share in disclosure's session (\% facts) & 14.3 & 7.0 & 9.6 & 20.2 & 20.4 \\
\GrayRule
\SecRow{6}{Recovery}
\FaintRule
Minimal sufficient sets & 15.7 & 16.0 & 17.2 & 15.0 & 14.6 \\
Turns per set (none of size 1) & 2.7 & 2.4 & 3.0 & 3.0 & 2.6 \\
\GrayRule
\SecRow{6}{Deletion}
\FaintRule
Minimal cut, mean & 3.04 & 3.22 & \cellcolor{leak!24}\textbf{2.68} & 3.08 & 3.16 \\
Minimal cut, max (budget: 10) & 5 & 5 & 4 & 5 & 4 \\
Cuttable with 2 deletions (\%) & 25.9 & 27.8 & \cellcolor{leak!24}\textbf{44.7} & 17.5 & 13.3 \\
\GrayRule
\SecRow{6}{Distraction}
\FaintRule
High-confusion distractors per fact & 26.6 & 27.3 & 26.4 & 27.3 & 25.5 \\
\BotRule
\end{tabular}
\caption{Per-fact statistics of \memcut{}, overall and by derivation type (means unless noted). \emph{Share in disclosure's session}: facts with at least one evidence turn in the disclosure's session. Sufficient sets and minimal cuts are author-identified and cross-checked by two independent algorithms. Every fact can be cut with at most five deletions, half the ten-turn budget; relation chains (highlighted) need the fewest. No evidence turn is shared across facts, and 62.6\% of turns carry no annotation.}
\label{tab:claims}
\end{table*}

\begin{table}[t]
\centering
\footnotesize
\TabSetup
\setlength{\tabcolsep}{2.8pt}
\renewcommand{\arraystretch}{1.0}
\begin{tabular}{@{}lVrrrrrrr@{}}
\TopRule
\textbf{EP} & Turns & Tokens & Sess. & \makecell[r]{Tok./\\turn} & \makecell[r]{Turns/\\sess.} & Days & Facts \\
\GrayRule
EP01 & 2,136 & 103,805 & 26 & 48.6 & 82.2 & 30 & 28 \\
EP02 & 2,152 & 77,846 & 28 & 36.2 & 76.9 & 27 & 28 \\
EP03 & 2,200 & 85,815 & 28 & 39.0 & 78.6 & 27 & 28 \\
EP04 & 2,200 & 70,715 & 29 & 32.1 & 75.9 & 28 & 27 \\
EP05 & 2,738 & 99,776 & 33 & 36.4 & 83.0 & 32 & 33 \\
EP06 & 2,798 & 100,768 & 35 & 36.0 & 79.9 & 34 & 32 \\
EP07 & 3,156 & 115,627 & 35 & 36.6 & 90.2 & 34 & 32 \\
EP08 & 3,278 & 101,697 & 32 & 31.0 & 102.4 & 31 & 32 \\
EP09 & 3,374 & 125,158 & 34 & 37.1 & 99.2 & 33 & 32 \\
EP10 & 3,302 & 143,060 & 35 & 43.3 & 94.3 & 34 & 32 \\
EP11 & 2,854 & 114,192 & 34 & 40.0 & 83.9 & 33 & 28 \\
EP12 & 3,258 & 157,627 & 36 & 48.4 & 90.5 & 81 & 32 \\
EP13 & 3,232 & 158,944 & 35 & 49.2 & 92.3 & 72 & 30 \\
EP14 & 3,472 & 169,198 & 36 & 48.7 & 96.4 & 86 & 32 \\
EP15 & 3,252 & 115,771 & 34 & 35.6 & 95.7 & 77 & 30 \\
\GrayRule
\rowcolor{tabhl}\textbf{Mean} & 2,893 & 116,000 & 32.7 & 39.9 & 88.1 & 43.9 & 30.4 \\
\BotRule
\end{tabular}
\caption{Per-episode statistics: turns, tokens (\texttt{o200k\_base}), sessions, tokens per turn, turns per session, days spanned, and protected facts.}
\label{tab:eps}
\end{table}

\paragraph{Derivation types.}
The four types abstract the most typical ways separate records entail a proposition. \emph{Fact assembly} joins attributes of one entity: a code name, a site visit, and an address make an acquisition target. \emph{Relation chains} join entities: a contract names a firm, a filing names its owner, a dinner names the owner's spouse, and a staff list names the spouse's role. \emph{Aggregation} combines values: several payments sum to a debt, several visits make a pattern. \emph{Temporal and state reasoning} orders records in time: an offer, a revision, and a withdrawal fix what holds now. Longer derivations often compose them.

\paragraph{Scale and balance.}
The facts are balanced across types, and every episode is long. The 456 protected facts split evenly over the four derivation types, 113--115 each (Table~\ref{tab:claims}), and every type appears in every episode six to nine times. The fifteen episodes that carry them hold 43,402 turns in 490 sessions (1.74M tokens); each averages 2,893 turns, 116k tokens, and 33 sessions over 27--86 days (Table~\ref{tab:eps}).

\paragraph{Derivation structure.}
Every fact can be derived in many ways, but never from a single evidence turn. A fact has 8.1 evidence turns on average (6--10; 3,700 in all), which the authors grouped into 7,148 minimal sufficient sets, 15.7 per fact (4--38), each of two to five turns (mean 2.7). Breaking all of them takes two deletions for 25.9\% of facts, three for 45.4\%, four for 28.1\%, and five for 0.7\%; the minimal cuts were computed by enumeration and agree with an independent dynamic program on all 456 facts. Each fact also has 26.6 distractors (16--88; 12,092 distinct turns). No evidence turn serves two facts, five turns are evidence for one fact and a distractor for another, and 62.6\% of turns carry no annotation.

\paragraph{Dispersion.}
Evidence is spread across the episode and mostly precedes the disclosure. A fact's evidence spans 8.0 sessions (5--10), with a median first-to-last span of 2,259 turns and 29 days, and lies a median of 714 turns and eight sessions from the disclosure. Only 67 of the 3,700 evidence turns (1.8\%) share the disclosure's session, and 14.3\% of facts have at least one such turn. Disclosures fall in the last third of the episode for 61\% of facts. Time is annotated at the granularity of session dates.

\section{Extended Results}
\label{app:extended}

This appendix gives the complete results behind \S\ref{sec:results} and \S\ref{sec:analysis}: recovery at every deletion level (\S\ref{app:levels}), the retrieval unit and derivation (\S\ref{app:unit}), and where cleanup loses the evidence (\S\ref{app:reach}).

\begin{table}[t]
\centering
\scriptsize
\TabSetup
\renewcommand{\SecStrut}{\rule[-3.3pt]{0pt}{10.8pt}}\renewcommand{\HeadTop}{\rule{0pt}{8.2pt}}\renewcommand{\HeadBot}{\rule[-3.6pt]{0pt}{0pt}}
\setlength{\tabcolsep}{2.6pt}
\renewcommand{\arraystretch}{1.12}
\begin{tabular}{@{}lV*{2}{c}V*{2}{c}V*{2}{c}V*{2}{c}@{}}
\TopRule
\HeadTop & \multicolumn{2}{cV}{\Head{DeepSeek}} & \multicolumn{2}{cV}{\Head{Qwen}} & \multicolumn{2}{cV}{\Head{GLM}} & \multicolumn{2}{c}{\Head{MiMo}} \\
\arrayrulecolor{tabblue}\cmidrule(lr){2-3}\cmidrule(lr){4-5}\cmidrule(lr){6-7}\cmidrule(l){8-9}\arrayrulecolor{tabgray}
\textbf{System}\HeadBot & \Sub{Nat.} & \Sub{Res.} & \Sub{Nat.} & \Sub{Res.} & \Sub{Nat.} & \Sub{Res.} & \Sub{Nat.} & \Sub{Res.} \\
\GrayRule
\SecRow{9}{Retrieval stores}
\FaintRule
Turn RAG & 95.0 & 100.0 & 90.8 & 100.0 & 92.1 & 100.0 & 92.8 & 100.0 \\
Session RAG & 62.5 & 100.0 & 48.2 & 100.0 & 52.9 & 100.0 & 69.7 & 100.0 \\
\GrayRule
\SecRow{9}{Memory systems}
\FaintRule
MemoryBank & 89.9 & 99.8 & 89.0 & 99.8 & 90.6 & 99.8 & 89.5 & 99.3 \\
Nemori & 95.2 & 98.2 & 89.0 & 97.4 & 93.9 & 93.9 & 95.8 & 99.1 \\
LightMem & 86.2 & 98.0 & 82.7 & 96.1 & 89.0 & 97.6 & 82.5 & 94.5 \\
\BotRule
\end{tabular}
\caption{Recovery (\%) before deletion. Every store supports nearly every fact (\emph{Res.}), and retrieval reveals most of them (\emph{Nat.}) in every system but Session RAG.}
\label{tab:full}
\end{table}

\begin{table*}[t]
\centering
\small
\TabSetup
\renewcommand{\SecStrut}{\rule[-4.3pt]{0pt}{13.8pt}}\renewcommand{\HeadTop}{\rule{0pt}{10.4pt}}\renewcommand{\HeadBot}{\rule[-4.6pt]{0pt}{0pt}}
\setlength{\tabcolsep}{3.4pt}
\renewcommand{\arraystretch}{1.08}
\begin{tabular}{@{}lV*{4}{c}V*{4}{c}V*{4}{c}V*{4}{c}@{}}
\TopRule
\HeadTop & \multicolumn{4}{cV}{\Head{DeepSeek}} & \multicolumn{4}{cV}{\Head{Qwen}} & \multicolumn{4}{cV}{\Head{GLM}} & \multicolumn{4}{c}{\Head{MiMo}} \\
\arrayrulecolor{tabblue}\cmidrule(lr){2-5}\cmidrule(lr){6-9}\cmidrule(lr){10-13}\cmidrule(l){14-17}\arrayrulecolor{tabgray}
\textbf{System}\HeadBot & \Sub{Full} & \Sub{S0} & \Sub{S1} & \Sub{S2} & \Sub{Full} & \Sub{S0} & \Sub{S1} & \Sub{S2} & \Sub{Full} & \Sub{S0} & \Sub{S1} & \Sub{S2} & \Sub{Full} & \Sub{S0} & \Sub{S1} & \Sub{S2} \\
\GrayRule
\SecRow{17}{Retrieval stores}
\FaintRule
Turn RAG & \cellcolor{leak!2}5.0 & \cellcolor{leak!12}39.1 & \cellcolor{leak!19}62.4 & \cellcolor{leak!23}76.7 & \cellcolor{leak!3}9.2 & \cellcolor{leak!13}42.2 & \cellcolor{leak!20}67.4 & \cellcolor{leak!25}82.0 & \cellcolor{leak!2}7.9 & \cellcolor{leak!12}39.5 & \cellcolor{leak!20}65.6 & \cellcolor{leak!24}78.5 & \cellcolor{leak!2}7.2 & \cellcolor{leak!12}38.7 & \cellcolor{leak!19}63.7 & \cellcolor{leak!23}76.4 \\
Session RAG & \cellcolor{leak!11}37.5 & \cellcolor{leak!23}75.3 & \cellcolor{leak!23}77.6 & \cellcolor{leak!25}82.6 & \cellcolor{leak!16}51.8 & \cellcolor{leak!24}79.7 & \cellcolor{leak!23}78.1 & \cellcolor{leak!25}83.3 & \cellcolor{leak!14}47.1 & \cellcolor{leak!25}82.4 & \cellcolor{leak!25}84.2 & \cellcolor{leak!28}92.3 & \cellcolor{leak!9}30.3 & \cellcolor{leak!21}71.0 & \cellcolor{leak!26}86.1 & \cellcolor{leak!26}85.4 \\
\GrayRule
\SecRow{17}{Memory systems}
\FaintRule
MemoryBank & \cellcolor{leak!3}9.9 & \cellcolor{leak!17}56.8 & \cellcolor{leak!21}71.5 & \cellcolor{leak!23}78.0 & \cellcolor{leak!3}10.8 & \cellcolor{leak!17}56.3 & \cellcolor{leak!22}73.0 & \cellcolor{leak!24}81.6 & \cellcolor{leak!3}9.2 & \cellcolor{leak!17}55.6 & \cellcolor{leak!22}73.8 & \cellcolor{leak!25}81.9 & \cellcolor{leak!3}9.9 & \cellcolor{leak!16}54.8 & \cellcolor{leak!21}71.3 & \cellcolor{leak!23}78.1 \\
Nemori & \cellcolor{leak!1}3.1 & \cellcolor{leak!9}30.2 & \cellcolor{leak!11}37.9 & \cellcolor{leak!13}42.7 & \cellcolor{leak!3}8.6 & \cellcolor{leak!13}41.9 & \cellcolor{leak!16}52.4 & \cellcolor{leak!17}55.9 & \cellcolor{leak!0}0.0 & \cellcolor{leak!9}30.8 & \cellcolor{leak!10}32.2 & \cellcolor{leak!12}40.6 & \cellcolor{leak!1}3.3 & \cellcolor{leak!10}33.6 & \cellcolor{leak!13}43.5 & \cellcolor{leak!14}47.8 \\
LightMem & \cellcolor{leak!4}12.1 & \cellcolor{leak!8}25.7 & \cellcolor{leak!11}37.1 & \cellcolor{leak!14}45.0 & \cellcolor{leak!4}13.9 & \cellcolor{leak!9}30.3 & \cellcolor{leak!14}45.2 & \cellcolor{leak!15}50.3 & \cellcolor{leak!3}8.8 & \cellcolor{leak!5}17.8 & \cellcolor{leak!11}35.6 & \cellcolor{leak!13}44.8 & \cellcolor{leak!4}12.8 & \cellcolor{leak!8}26.1 & \cellcolor{leak!11}35.8 & \cellcolor{leak!15}49.0 \\
\BotRule
\end{tabular}
\caption{Hidden-risk share (\%), computed as $(\mathrm{Res}-\mathrm{Nat})/\mathrm{Res}$ at every deletion level. It is a lower bound on the share of residual successes that native recovery misses (Appendix~\ref{app:levels}). Shading is proportional to the value. \emph{Full} is the store before any deletion. The share rises from Full to S2 in all 20 configurations.}
\label{tab:hr}
\end{table*}

\subsection{Recovery at every deletion level}
\label{app:levels}

\paragraph{Before deletion.}
Every system retains what recovery needs: residual recovery is 100\% in both verbatim stores and 94--100\% in the derived systems (Table~\ref{tab:full}), so later drops reflect deletion rather than encoding loss, apart from a small loss of aggregate values at write time in LightMem (Appendix~\ref{app:types}). Native recovery is 82--96\% in every system except Session RAG, where it is 48--70\% because three retrieved sessions hold only part of a fact's evidence (\S\ref{app:unit}).

\paragraph{Write models.}
Within each system the four write models agree on residual recovery after S2 to within about five points, and to within ten for Session RAG. The main patterns of \S\ref{sec:results} are consistent across these four models within each system.

\paragraph{Hidden-risk share.}
The hidden-risk share rises from Full to S2 in all 20 configurations, monotonically in the 16 outside Session RAG (Table~\ref{tab:hr}). Session RAG hides the most at every level, from 30--52\% before any deletion to 83--92\% after S2.
The share is a conservative estimate.
Native and residual recovery are answered independently.
As a result, native recovery finds a few facts that residual recovery misses.
In the six configurations that we analyze per fact (MemoryBank, Nemori, and LightMem with DeepSeek and Qwen), these facts make up 4.8\% of native successes.
Let $R$ and $N$ be the sets of facts that residual and native recovery find.
The share of facts in $R$ that native recovery misses is $|R\setminus N|/|R|$.
The reported share also subtracts the facts in $N$ that are not in $R$, so it is a lower bound on $|R\setminus N|/|R|$.
In the six configurations, $|R\setminus N|/|R|$ is 0--12 points higher than the reported share, and the gap is largest after S2.
After S2 it is 80--85\% in MemoryBank and 52--62\% in Nemori and LightMem, while the reported share is 78--82\% and 43--56\%.
It also rises from Full to S2 in all six configurations.
The reported share is therefore conservative about the level of hidden risk in these configurations.

\paragraph{Collateral deletion at S0.}
Most of the residual recovery that derived systems lose at S0 was carried away by closure deletion: a derived object that also contains the disclosure goes with it, together with any evidence it summarized. The largest overall loss, 13 points for Nemori with DeepSeek, is close to the share of facts with at least one evidence turn in the disclosure's session (14.3\%, Table~\ref{tab:claims}). Four fifths of the loss falls on aggregation, for reasons given in Appendix~\ref{app:types}.

\subsection{Retrieval unit and derivation strength}
\label{app:unit}

\paragraph{Retrieval unit.}
Session retrieval reads more and finds less. At the evaluated retrieval depths, a Session RAG context averages 13.3k tokens and holds 1.9 evidence turns, a Turn RAG context 1.1k tokens and 4.8. After the disclosure is deleted, the two stores' residual recovery agrees to within about two points in 15 of the 16 model--type cells of Table~\ref{tab:rag}, while Session RAG's native recovery is 31--43 points lower and its hidden-risk share is 71--82\% against 39--42\%. After cleanup Session RAG also leaves 10--23 points more recoverable, because its candidate pools are diluted by non-evidence turns from the same sessions.

\paragraph{Derivation strength.}
Retrieval units help explain native recovery. In MemoryBank it is one turn, so once the disclosure is gone each evidence turn has to match the question on its own: deleting the disclosure costs 46--50 points of native recovery, and after cleanup four in five of the 72--76\% of facts that remain recoverable are unseen by the system's own retrieval. In Nemori it is a narrative or a distilled fact that fuses several turns, and in LightMem a consolidated entry that fuses more; such objects can still match the question without the disclosure, consistent with the smaller losses of 29--36 and 12--20 points. What a system retains follows instead how much evidence its deletion pool reaches (\S\ref{app:reach}): MemoryBank's pool of twenty holds 3.0 of a fact's evidence turns against 4.5 in Turn RAG and LightMem, and cleanup lowers its residual recovery by only 18--23 points, against 33--38 in Nemori and 47--52 in LightMem. Closure deletion also removes a whole object when any of its source turns goes; what it costs in unrelated memory is a separate question, which we leave to future work.

\subsection{Where cleanup loses the evidence}
\label{app:reach}

\begin{table*}[t]
\centering
\footnotesize
\TabSetup
\renewcommand{\SecStrut}{\rule[-3.9pt]{0pt}{12.6pt}}\renewcommand{\HeadTop}{\rule{0pt}{9.6pt}}\renewcommand{\HeadBot}{\rule[-4.2pt]{0pt}{0pt}}
\setlength{\tabcolsep}{3.3pt}
\renewcommand{\arraystretch}{1.06}
\begin{tabular*}{\textwidth}{@{\extracolsep{\fill}}lV*{4}{c}V*{4}{c}V*{4}{c}@{}}
\TopRule
\HeadTop & \multicolumn{4}{cV}{\Head{Turn RAG}} & \multicolumn{4}{cV}{\Head{MemoryBank}} & \multicolumn{4}{c}{\Head{LightMem}} \\
\arrayrulecolor{tabblue}\cmidrule(lr){2-5}\cmidrule(lr){6-9}\cmidrule(l){10-13}\arrayrulecolor{tabgray}
 & \multicolumn{2}{c}{DeepSeek} & \multicolumn{2}{cV}{Qwen} & \multicolumn{2}{c}{DeepSeek} & \multicolumn{2}{cV}{Qwen} & \multicolumn{2}{c}{DeepSeek} & \multicolumn{2}{c}{Qwen} \\
\cmidrule(lr){2-3}\cmidrule(lr){4-5}\cmidrule(lr){6-7}\cmidrule(lr){8-9}\cmidrule(lr){10-11}\cmidrule(l){12-13}
\HeadBot & \Sub{S1} & \Sub{S2} & \Sub{S1} & \Sub{S2} & \Sub{S1} & \Sub{S2} & \Sub{S1} & \Sub{S2} & \Sub{S1} & \Sub{S2} & \Sub{S1} & \Sub{S2} \\
\GrayRule
\SecRow{13}{Widening the candidate pool \textmd{\textcolor{black!55}{(budget $=10$; first column: pool size)}}}
\FaintRule
10 & {\color{black!62}66.7} & \cellcolor{leak!20}66.7 & {\color{black!62}63.2} & \cellcolor{leak!19}62.9 & {\color{black!62}80.6} & \cellcolor{leak!24}80.6 & {\color{black!62}80.6} & \cellcolor{leak!24}80.6 & {\color{black!62}63.2} & \cellcolor{leak!19}63.4 & {\color{black!62}57.6} & \cellcolor{leak!17}56.1 \\
\rowcolor{tabhl}\textbf{20}$^{\star}$ & {\color{black!62}66.7} & \cellcolor{leak!16}52.1 & {\color{black!62}63.2} & \cellcolor{leak!14}47.9 & {\color{black!62}80.6} & \cellcolor{leak!22}72.2 & {\color{black!62}80.6} & \cellcolor{leak!22}72.2 & {\color{black!62}62.5} & \cellcolor{leak!15}49.3 & {\color{black!62}52.1} & \cellcolor{leak!14}47.2 \\
30 & {\color{black!62}66.7} & \cellcolor{leak!14}47.2 & {\color{black!62}63.2} & \cellcolor{leak!13}44.1 & {\color{black!62}80.6} & \cellcolor{leak!20}68.1 & {\color{black!62}80.6} & \cellcolor{leak!18}58.7 & {\color{black!62}62.5} & \cellcolor{leak!13}44.9 & {\color{black!62}52.1} & \cellcolor{leak!12}39.3 \\
40 & {\color{black!62}66.7} & \cellcolor{leak!13}43.8 & {\color{black!62}63.2} & \cellcolor{leak!12}39.2 & {\color{black!62}80.6} & \cellcolor{leak!18}61.5 & {\color{black!62}80.6} & \cellcolor{leak!18}60.0 & {\color{black!62}62.5} & \cellcolor{leak!11}37.9 & {\color{black!62}52.1} & \cellcolor{leak!11}37.0 \\
50 & {\color{black!62}66.7} & \cellcolor{leak!12}40.3 & {\color{black!62}63.2} & \cellcolor{leak!11}35.5 & {\color{black!62}80.6} & \cellcolor{leak!19}62.5 & {\color{black!62}80.6} & \cellcolor{leak!16}51.9 & {\color{black!62}62.5} & \cellcolor{leak!12}39.4 & {\color{black!62}52.1} & \cellcolor{leak!10}33.8 \\
60 & {\color{black!62}66.7} & \cellcolor{leak!12}39.6 & {\color{black!62}63.2} & \cellcolor{leak!9}30.8 & {\color{black!62}80.6} & \cellcolor{leak!17}58.0 & {\color{black!62}80.6} & \cellcolor{leak!14}48.3 & {\color{black!62}62.5} & \cellcolor{leak!10}34.3 & {\color{black!62}52.1} & \cellcolor{leak!9}31.6 \\
\GrayRule
\SecRow{13}{Raising the deletion budget \textmd{\textcolor{black!55}{(pool $=20$; first column: budget)}}}
\FaintRule
2 & {\color{black!62}95.1} & \cellcolor{leak!28}93.1 & {\color{black!62}93.1} & \cellcolor{leak!27}90.9 & {\color{black!62}91.0} & \cellcolor{leak!27}91.0 & {\color{black!62}86.4} & \cellcolor{leak!27}90.7 & {\color{black!62}88.9} & \cellcolor{leak!26}86.1 & {\color{black!62}85.4} & \cellcolor{leak!25}81.9 \\
4 & {\color{black!62}84.0} & \cellcolor{leak!20}66.0 & {\color{black!62}81.8} & \cellcolor{leak!20}68.1 & {\color{black!62}91.0} & \cellcolor{leak!27}89.6 & {\color{black!62}87.3} & \cellcolor{leak!26}88.2 & {\color{black!62}81.9} & \cellcolor{leak!19}63.2 & {\color{black!62}78.5} & \cellcolor{leak!21}69.9 \\
6 & {\color{black!62}76.4} & \cellcolor{leak!16}54.9 & {\color{black!62}72.9} & \cellcolor{leak!16}52.1 & {\color{black!62}86.1} & \cellcolor{leak!24}80.6 & {\color{black!62}80.9} & \cellcolor{leak!23}76.6 & {\color{black!62}72.7} & \cellcolor{leak!16}52.8 & {\color{black!62}68.5} & \cellcolor{leak!16}54.9 \\
\rowcolor{tabhl}\textbf{10}$^{\star}$ & {\color{black!62}66.7} & \cellcolor{leak!16}52.1 & {\color{black!62}63.2} & \cellcolor{leak!14}47.9 & {\color{black!62}80.6} & \cellcolor{leak!22}72.2 & {\color{black!62}80.6} & \cellcolor{leak!22}72.2 & {\color{black!62}62.5} & \cellcolor{leak!15}49.3 & {\color{black!62}52.1} & \cellcolor{leak!14}47.2 \\
15 & {\color{black!62}54.2} & \cellcolor{leak!15}51.4 & {\color{black!62}51.8} & \cellcolor{leak!14}47.2 & {\color{black!62}73.6} & \cellcolor{leak!21}69.4 & {\color{black!62}75.4} & \cellcolor{leak!20}67.1 & {\color{black!62}53.5} & \cellcolor{leak!14}47.2 & {\color{black!62}45.1} & \cellcolor{leak!13}44.4 \\
20 & {\color{black!62}50.7} & \cellcolor{leak!15}50.7 & {\color{black!62}47.2} & \cellcolor{leak!14}47.2 & {\color{black!62}72.2} & \cellcolor{leak!22}72.2 & {\color{black!62}65.7} & \cellcolor{leak!20}65.7 & {\color{black!62}45.8} & \cellcolor{leak!14}45.8 & {\color{black!62}43.1} & \cellcolor{leak!13}43.1 \\
\BotRule
\end{tabular*}
\caption{Residual recovery (\%) behind Figure~\ref{fig:analysis}a--b: retrieval-ranked (S1) and LLM-guided (S2) cleanup as the candidate pool widens with the budget fixed at ten turns (top), and as the budget grows with the pool fixed at twenty (bottom). The sweeps run on the 144 facts of EP01--05 (Appendix~\ref{app:reach}), so values at the main-table setting ($^{\star}$) differ from Tables~\ref{tab:rag} and~\ref{tab:mem}, which cover all 456. S2 is shaded in proportion to its value. Percentages are over completed runs: 98.9\% of those planned, and at least 114 of the 144 facts in every cell.}
\label{tab:sweep}
\end{table*}

Table~\ref{tab:sweep} gives every point of Figure~\mbox{\ref{fig:analysis}a--b}. The sweeps cover the 144 facts of EP01--05, so values at the main setting differ from Tables~\ref{tab:rag} and~\ref{tab:mem}, and the sweeps are read as changes from their own main setting.

\paragraph{Candidate coverage.}
Candidate pools omit supporting evidence. At the main pool of 20, the candidates contain on average 4.5 of a fact's evidence turns in Turn RAG and LightMem and 3.0 in MemoryBank, and S2 deletes 4.4, 4.3, and 2.7 of them (means over the two write models). Widening the pool to 60 raises the evidence it contains to 5.8, 5.9, and 4.5, still fewer than the six to ten evidence turns every fact has (Appendix~\ref{app:data}), and S2 deletes 5.5, 5.2, and 3.4. Raising the budget cannot go beyond the pool: at a budget of 20 the deleter removes the entire pool, S1 and S2 coincide, and 43--51\% of facts remain recoverable in Turn RAG and LightMem and 66--72\% in MemoryBank. In LightMem some pools of 20 hold fewer deletable turns than the budget, so a budget of 20 deletes 19.1 turns on average with DeepSeek and 16.6 with Qwen rather than 20.

\paragraph{Evidence deleted on the full benchmark.}
The counts above come from the 144-fact sweep subset.
Table~\ref{tab:evdel} repeats the count of deleted evidence on all 456 facts in the three memory systems with DeepSeek and Qwen: of the ten turns that cleanup deletes for a fact, it reports how many are annotated evidence.
With ten additional selected turns, $k$ selected evidence turns imply an off-target fraction of $1-k/10$.

\paragraph{Evidence removed by cleanup.}
Of the ten cleanup selections, 3.0 are evidence under S1 and 3.6 under S2, corresponding to approximately 70\% and 64\% off-target selections. An average fact has 8.1 evidence turns (Table~\ref{tab:claims}).
This is consistent with the sweeps, where the pool of twenty holds only about half of a fact's evidence (\S\ref{sec:reach}).
LLM judgment adds 0.7 evidence turns per fact over retrieval ranking; the sweeps show larger recovery reductions from widening the pool than from deleting its remaining candidates.

\paragraph{Memory design changes reach.}
MemoryBank, whose pool holds the least evidence, deletes the least, 2.9 evidence turns per fact under S2, and keeps the most facts recoverable (72--73\% residual recovery after S2; Table~\ref{tab:mem}).
LightMem deletes the most, 4.7, and keeps the fewest (39--42\%), and Nemori lies between them on both (3.4; 52--57\%).
Across systems, residual recovery thus follows the evidence a system removes, as it does across the settings of the sweeps.

\paragraph{Relation chains are the hardest to reach.}
In every system and under both strategies, relation chains lose the fewest evidence turns: 2.4 per fact under S2, against 3.8 for fact assembly, 3.9 for temporal state, and 4.6 for aggregation.
The gap holds within each system, at 1.7 against 3.0--3.7 in MemoryBank, 2.4 against 3.5--3.9 in Nemori, and 2.9 against 4.8--6.0 in LightMem.
Their evidence names the request's people and firms but shares little of its wording, so similarity retrieval ranks it low (\S\ref{sec:types}).

\begin{table}[t]
\centering
\footnotesize
\setlength{\aboverulesep}{0.2ex}
\setlength{\belowrulesep}{0.3ex}
\renewcommand{\arraystretch}{0.95}
\newcommand{\EvBody}{%
\arrayrulecolor{accent}\specialrule{0.9pt}{0pt}{1pt}\arrayrulecolor{black}
\textbf{System} & & \textbf{Assem.} & \textcolor{accent}{\textbf{Chain}} & \textbf{Aggr.} & \textbf{Temp.} & \textbf{All} \\
\arrayrulecolor{accent!60}\midrule\arrayrulecolor{black}
MemoryBank & \textcolor{black!65}{S1} & 2.2 & \textcolor{accent}{1.2} & 2.9 & 2.3 & 2.2 \\
 & \textcolor{black!65}{S2} & 3.0 & \textcolor{accent}{1.7} & 3.7 & 3.2 & 2.9 \\
\addlinespace[1pt]
Nemori & \textcolor{black!65}{S1} & 2.9 & \textcolor{accent}{1.8} & 3.5 & 2.9 & 2.8 \\
 & \textcolor{black!65}{S2} & 3.6 & \textcolor{accent}{2.4} & 3.9 & 3.5 & 3.4 \\
\addlinespace[1pt]
LightMem & \textcolor{black!65}{S1} & 3.9 & \textcolor{accent}{2.4} & 5.2 & 4.2 & 3.9 \\
 & \textcolor{black!65}{S2} & 4.8 & \textcolor{accent}{2.9} & 6.0 & 5.1 & 4.7 \\
\arrayrulecolor{accent!60}\midrule\arrayrulecolor{black}
\textbf{All three} & \textcolor{black!65}{S1} & 3.0 & \textcolor{accent}{1.8} & 3.9 & 3.2 & 3.0 \\
 & \textcolor{black!65}{S2} & 3.8 & \textcolor{accent}{2.4} & 4.6 & 3.9 & 3.6 \\
\arrayrulecolor{accent}\specialrule{0.9pt}{0pt}{0pt}\arrayrulecolor{black}}%
\setbox0=\hbox{\setlength{\tabcolsep}{0pt}\begin{tabular}{ll*{5}{c}}\EvBody\end{tabular}}%
\setlength{\tabcolsep}{\dimexpr(\columnwidth-\wd0)/14-0.01pt\relax}%
\begin{tabular}{ll*{5}{c}}\EvBody\end{tabular}
\caption{Annotated evidence turns among the ten turns that cleanup deletes per fact, by system and derivation type (\emph{Assem.}: fact assembly; \emph{Chain}: relation chain; \emph{Aggr.}: aggregation; \emph{Temp.}: temporal state), over all 456 facts and the DeepSeek and Qwen write models.}
\label{tab:evdel}
\end{table}

\paragraph{Judgment grows with reach.}
S1 deletes the ten top-ranked candidates, so its deletion set is unchanged across the pool sweep once the pool holds ten deletable turns. At a pool of ten both strategies delete the same turns. The gain of S2 over S1 is therefore what judgment adds from candidates beyond the top ten: 5--15 points at a pool of 20 and 20--32 at a pool of 60. Judgment is worth what reach gives it to choose from.

\paragraph{Partial cuts leave derivations intact.}
A complete cut intersects every annotated sufficient set. At a pool of 60, S2 intersects 92--93\% of a fact's sufficient sets on average in Turn RAG and LightMem; residual recovery is still 31--40\%.

\paragraph{Types.}
Widening reaches more aggregation evidence than relation-chain evidence. At a pool of 60 the Turn RAG candidates contain 6.9 of an aggregate's evidence turns but 4.4 of a relation chain's, and after S2, 3--31\% of aggregates and 49--78\% of relation chains remain recoverable across the three systems and two write models. Relation chains have higher residual recovery than aggregation throughout both sweeps, the pattern of Appendix~\ref{app:types}.

\paragraph{One quantity behind both sweeps.}
Across the 132 distinct settings of the two sweeps (both strategies, three systems, two write models), residual recovery falls almost linearly with the mean number of annotated evidence turns a setting removes per fact, by about 13 points per turn ($R^2=0.95$). Given that number, the system and the write model move residual recovery by at most about four points, and the strategy by less than one. The effects of pool width, budget, and judgment on residual recovery are closely associated with how much evidence they remove, consistent with MemoryBank retaining more because its pool holds less evidence. The fitted line reaches zero near 8.2 removed turns, close to the 8.1 evidence turns of an average fact (Table~\ref{tab:claims}).

\subsection{Deletion cost and residual recovery}
\label{app:cost}

We compare recovery at specified source-turn deletion budgets. Selecting the entire source history exceeds these budgets. The smallest annotated cut contains 3.04 evidence turns on average and at most five (Table~\ref{tab:claims}). Forgetting therefore need not require deleting every evidence turn.

For the additional selected turns $A$ and the target's evidence $E$, the off-target fraction is \mbox{$|A\setminus E|/|A|$}. The denominator is the actual number selected, excluding the common disclosure removal. This is a source-level reference measure of collateral damage.

\begin{table}[t]
\centering
\footnotesize
\TabSetup
\renewcommand{\arraystretch}{1.1}
\setlength{\tabcolsep}{4pt}
\begin{tabular*}{\columnwidth}{@{\extracolsep{\fill}}rrrr@{}}
\TopRule
\Head{Pool} & \Head{Selected} & \Head{Off-target} & \Head{Res. (\%)} \\
\GrayRule
20 & 10 & 5.54 & 52.08 \\
20 & 20 & 15.49 & 50.69 \\
60 & 10 & 4.50 & 39.58 \\
\BotRule
\end{tabular*}
\caption{Deletion cost and residual recovery for DeepSeek--Turn RAG, S2, on the same 144 facts, all completed. Pool, selected, and off-target counts are in source turns. Selected and off-target counts exclude disclosure removal and are means per fact.}
\label{tab:deletion_cost}
\end{table}

Table~\ref{tab:deletion_cost} reuses three settings of the existing sweeps on the same 144 facts. Turn RAG stores one object per turn, so closure adds no other records. Doubling the budget from ten to twenty adds 9.95 non-evidence deletions per fact on average, while residual recovery falls by only 1.39 points. Widening the pool from twenty to sixty at budget ten instead lowers residual recovery by 12.50 points and removes 1.04 fewer non-evidence turns. Thus, broader coverage improves forgetting without selecting more turns and reduces off-target deletion in this comparison.

These results do not establish downstream utility preservation. Evidence turns can serve legitimate tasks, and deleting evidence beyond a complete cut can also be unnecessary. In derived stores, closure may remove unrelated content from shared objects, which the source-turn budget does not bound. Recovery and source-level deletion cost should therefore be read together, while retained task performance remains a separate evaluation dimension.

\begin{table}[t]
\centering
\footnotesize
\TabSetup
\renewcommand{\SecStrut}{\rule[-3.9pt]{0pt}{12.6pt}}\renewcommand{\HeadTop}{\rule{0pt}{9.6pt}}\renewcommand{\HeadBot}{\rule[-4.2pt]{0pt}{0pt}}
\setlength{\tabcolsep}{3.2pt}
\renewcommand{\arraystretch}{1.04}
\begin{tabular*}{\columnwidth}{@{\extracolsep{\fill}}lVcccc@{}}
\TopRule
\HeadTop & \multicolumn{4}{c}{\Head{Derivation type}} \\
\arrayrulecolor{tabblue}\cmidrule(l){2-5}\arrayrulecolor{tabgray}
\HeadBot & Fact & Rel. & Aggr. & Temp. \\
\GrayRule
\SecRow{5}{Recovery, mean \%}
\FaintRule
Native before deletion & 91.9 & \textbf{81.1} & 90.4 & 96.1 \\
Native after S0 (visible) & 73.8 & \textbf{18.2} & 60.9 & 76.6 \\
Residual lost at S0 & $-$0.1 & 2.3 & \textbf{14.7} & 0.9 \\
Residual recovery after S2 & 60.2 & \textbf{71.8} & 24.3 & 58.4 \\
Removed by cleanup, S0$\to$S2 & 38.9 & 24.1 & \textbf{57.7} & 40.1 \\
Hidden-risk share after S2 & 49.2 & \textbf{87.1} & 66.3 & 52.8 \\
Gain from LLM judgment & 14.3 & \textbf{8.3} & 14.3 & 11.0 \\
\GrayRule
\SecRow{5}{Annotated structure}
\FaintRule
Minimal cut (turns) & 3.22 & \textbf{2.68} & 3.08 & 3.16 \\
Cut by two deletions (\%) & 27.8 & \textbf{44.7} & 17.5 & 13.3 \\
\BotRule
\end{tabular*}
\caption{Leakage by derivation type: means over the 16 configurations outside Session RAG (four write models, four systems), with the hidden-risk share computed from the mean recovery values. \emph{LLM judgment}: residual recovery lowered by S2 relative to S1. Annotated structure from Table~\ref{tab:claims}. Fact: fact assembly; Rel.: relation chain; Aggr.: aggregation; Temp.: temporal state. Bold marks the type that stands out in each row.}
\label{tab:types}
\end{table}

\begin{table*}[t]
\centering
\footnotesize
\TabSetup
\renewcommand{\SecStrut}{\rule[-3.9pt]{0pt}{12.6pt}}\renewcommand{\HeadTop}{\rule{0pt}{9.6pt}}\renewcommand{\HeadBot}{\rule[-4.2pt]{0pt}{0pt}}
\setlength{\tabcolsep}{4pt}
\renewcommand{\arraystretch}{1.0}
\begin{tabular*}{\textwidth}{@{\extracolsep{\fill}}llV*{2}{c}V*{2}{c}V*{2}{c}V*{2}{c}@{}}
\TopRule
\HeadTop & & \multicolumn{2}{cV}{\Head{Full}} & \multicolumn{2}{cV}{\Head{S0}} & \multicolumn{2}{cV}{\Head{S1}} & \multicolumn{2}{c}{\Head{S2}} \\
\arrayrulecolor{tabblue}\cmidrule(lr){3-4}\cmidrule(lr){5-6}\cmidrule(lr){7-8}\cmidrule(l){9-10}\arrayrulecolor{tabgray}
\textbf{Type}\HeadBot & \textbf{System} & \Sub{Nat.} & \Sub{Res.} & \Sub{Nat.} & \Sub{Res.} & \Sub{Nat.} & \Sub{Res.} & \Sub{Nat.} & \Sub{Res.} \\
\SecRow{10}{DeepSeek}
\FaintRule
\multirow{3}{*}{Fact assembly} & MemoryBank & 85.2 & \cellcolor{leak!30}100.0 & 53.0 & \cellcolor{leak!30}100.0 & 32.2 & \cellcolor{leak!28}92.2 & 23.5 & \cellcolor{leak!25}84.3 \\
& Nemori & 94.8 & \cellcolor{leak!29}98.3 & 81.7 & \cellcolor{leak!29}97.4 & 65.2 & \cellcolor{leak!25}84.3 & 50.4 & \cellcolor{leak!19}63.5 \\
& LightMem & 97.4 & \cellcolor{leak!30}100.0 & 93.0 & \cellcolor{leak!30}99.1 & 53.9 & \cellcolor{leak!19}62.6 & 40.0 & \cellcolor{leak!15}51.3 \\
\FaintRule
\multirow{3}{*}{Relation chain} & MemoryBank & 84.2 & \cellcolor{leak!30}100.0 & 7.0 & \cellcolor{leak!29}95.6 & 5.3 & \cellcolor{leak!27}88.6 & 6.1 & \cellcolor{leak!26}86.8 \\
& Nemori & 93.0 & \cellcolor{leak!29}98.2 & 17.5 & \cellcolor{leak!27}89.5 & 18.4 & \cellcolor{leak!22}73.7 & 11.4 & \cellcolor{leak!18}60.5 \\
& LightMem & 74.6 & \cellcolor{leak!29}96.5 & 27.2 & \cellcolor{leak!28}93.9 & 22.8 & \cellcolor{leak!23}77.2 & 15.8 & \cellcolor{leak!21}70.2 \\
\FaintRule
\multirow{3}{*}{Aggregation} & MemoryBank & 93.9 & \cellcolor{leak!30}99.1 & 45.6 & \cellcolor{leak!24}79.8 & 20.2 & \cellcolor{leak!18}58.8 & 7.0 & \cellcolor{leak!11}37.7 \\
& Nemori & 93.0 & \cellcolor{leak!29}98.2 & 59.6 & \cellcolor{leak!18}58.8 & 21.1 & \cellcolor{leak!10}32.5 & 14.9 & \cellcolor{leak!6}21.1 \\
& LightMem & 79.8 & \cellcolor{leak!29}95.6 & 55.3 & \cellcolor{leak!19}64.0 & 12.3 & \cellcolor{leak!6}20.2 & 6.1 & \cellcolor{leak!2}6.1 \\
\FaintRule
\multirow{3}{*}{Temporal state} & MemoryBank & 96.5 & \cellcolor{leak!30}100.0 & 54.9 & \cellcolor{leak!29}96.5 & 34.5 & \cellcolor{leak!25}84.1 & 26.5 & \cellcolor{leak!23}77.9 \\
& Nemori & 100.0 & \cellcolor{leak!29}98.2 & 78.8 & \cellcolor{leak!28}94.7 & 61.9 & \cellcolor{leak!23}77.9 & 43.4 & \cellcolor{leak!19}64.6 \\
& LightMem & 92.9 & \cellcolor{leak!30}100.0 & 88.5 & \cellcolor{leak!29}98.2 & 49.6 & \cellcolor{leak!18}60.2 & 30.1 & \cellcolor{leak!12}39.8 \\
\GrayRule
\SecRow{10}{Qwen}
\FaintRule
\multirow{3}{*}{Fact assembly} & MemoryBank & 81.7 & \cellcolor{leak!30}100.0 & 53.9 & \cellcolor{leak!29}97.4 & 28.7 & \cellcolor{leak!27}88.7 & 19.1 & \cellcolor{leak!25}83.5 \\
& Nemori & 86.1 & \cellcolor{leak!29}96.5 & 75.7 & \cellcolor{leak!30}99.1 & 54.8 & \cellcolor{leak!25}83.5 & 45.2 & \cellcolor{leak!20}67.0 \\
& LightMem & 95.7 & \cellcolor{leak!30}100.0 & 89.6 & \cellcolor{leak!30}100.0 & 48.7 & \cellcolor{leak!17}58.3 & 35.7 & \cellcolor{leak!14}47.0 \\
\FaintRule
\multirow{3}{*}{Relation chain} & MemoryBank & 84.2 & \cellcolor{leak!30}100.0 & 7.9 & \cellcolor{leak!29}96.5 & 5.3 & \cellcolor{leak!28}92.1 & 4.4 & \cellcolor{leak!25}84.2 \\
& Nemori & 90.4 & \cellcolor{leak!29}98.2 & 13.2 & \cellcolor{leak!27}90.4 & 8.8 & \cellcolor{leak!23}78.1 & 9.6 & \cellcolor{leak!19}64.9 \\
& LightMem & 66.7 & \cellcolor{leak!29}97.4 & 23.7 & \cellcolor{leak!28}94.7 & 11.4 & \cellcolor{leak!20}67.5 & 7.0 & \cellcolor{leak!17}57.0 \\
\FaintRule
\multirow{3}{*}{Aggregation} & MemoryBank & 96.5 & \cellcolor{leak!30}99.1 & 50.9 & \cellcolor{leak!27}89.5 & 17.5 & \cellcolor{leak!17}57.9 & 7.9 & \cellcolor{leak!15}50.9 \\
& Nemori & 87.7 & \cellcolor{leak!29}96.5 & 54.4 & \cellcolor{leak!24}81.6 & 21.9 & \cellcolor{leak!12}38.6 & 15.8 & \cellcolor{leak!9}30.7 \\
& LightMem & 75.4 & \cellcolor{leak!26}87.7 & 55.3 & \cellcolor{leak!20}67.5 & 5.3 & \cellcolor{leak!4}12.3 & 4.4 & \cellcolor{leak!2}7.9 \\
\FaintRule
\multirow{3}{*}{Temporal state} & MemoryBank & 93.8 & \cellcolor{leak!30}100.0 & 54.9 & \cellcolor{leak!30}100.0 & 35.4 & \cellcolor{leak!25}83.2 & 22.1 & \cellcolor{leak!22}71.7 \\
& Nemori & 92.0 & \cellcolor{leak!29}98.2 & 69.9 & \cellcolor{leak!29}95.6 & 45.1 & \cellcolor{leak!22}74.3 & 30.1 & \cellcolor{leak!20}66.4 \\
& LightMem & 92.9 & \cellcolor{leak!30}99.1 & 83.2 & \cellcolor{leak!30}99.1 & 38.9 & \cellcolor{leak!16}52.2 & 31.0 & \cellcolor{leak!14}45.1 \\
\BotRule
\end{tabular*}
\caption{Recovery (\%) by derivation type in the three memory systems with DeepSeek and Qwen, adding the pre-deletion baseline (Full) that Table~\ref{tab:mem} omits (113--115 facts per type). \emph{Res.} is shaded in proportion to its value, as in Table~\ref{tab:rag}.}
\label{tab:catfull}
\end{table*}

\section{Leakage by Derivation Type}
\label{app:types}

The four derivation types of \S\ref{sec:data} differ in how their evidence relates to a deletion request, not in how much evidence they have (Table~\ref{tab:claims}). Section~\ref{sec:analysis} reads them through derivation and reach (Figure~\mbox{\ref{fig:analysis}c--d}); Table~\ref{tab:types} summarizes each type over the 16 configurations outside Session RAG and Table~\ref{tab:catfull} gives the per-type results in full. After cleanup, relation chains have the highest residual recovery in 12 of these configurations and aggregation the lowest in all 16; temporal facts, whose time cues the question repeats, and fact assembly, whose evidence shares the request's wording, stay the most visible after S0 (77\% and 74\%). Four further regularities follow from the same relation between a type's evidence and the request.

\paragraph{Relation chains are hard to recall and forget.}
Relation chains are also the type retrieval answers least well before any deletion, 81\% against 90--96\% for the others and 68\% in LightMem, and deleting the disclosure costs them 63 points of native recovery against 18--30. Their visibility rests almost entirely on the one turn that states them: at S0, LightMem still answers 94\% of fact-assembly and 90\% of temporal questions but only 30\% of relation chains. Their evidence names the right people and firms but not in the words of the question or request, which may hinder similarity retrieval for both answering and deletion.

\paragraph{Judgment helps least where reach fails most.}
Replacing retrieval ranking with LLM judgment lowers residual recovery by 14 points for fact assembly and aggregation, 11 for temporal facts, and 8 for relation chains (Table~\ref{tab:types}). The smaller gain for relation chains is consistent with limited candidate coverage: judgment can select only turns already in the pool.

\paragraph{Coarse retrieval costs aggregation most.}
On the same text, Session RAG answers fewer questions than Turn RAG after S0 for every type, but by different margins: about 54 points for aggregation, 37 to 38 for fact assembly and temporal facts, and 17 for relation chains (Table~\ref{tab:rag}). An aggregate needs several values at once, and three retrieved sessions rarely contain all of them; relation chains lose least because turn retrieval already fails to surface them.

\paragraph{Derivation reveals every type but relation chains.}
For fact assembly, aggregation, and temporal facts, the hidden-risk share after S2 falls from 68--85\% in Turn RAG and MemoryBank to 20--39\% in Nemori and LightMem (Figure~\ref{fig:analysis}d), consistent with derived objects combining evidence within topics. Relation chains stay 81--94\% hidden in every system, likely because their links span topics (\S\ref{sec:design}).

\paragraph{Aggregation also loses what deletion did not target.}
Aggregation is the only type whose residual recovery falls when only the disclosure is deleted: 15 points on average against at most 3 for the others, and 4 to 13 points even in Turn RAG, which derives nothing. Co-location does not explain it, since temporal facts share a session with their disclosure as often, in a fifth of facts (Table~\ref{tab:claims}), and lose about one point. Two effects do. First, an aggregate has to be computed from several values, and the disclosure states the result, so once the disclosure is gone the reader has to carry out the computation itself, which it does less reliably than it reads a stated fact. The evidence still suffices by design. The reader recovers 96--99\% of facts from evidence alone in the validation of \S\ref{sec:data}, with misses concentrated in aggregation, consistent with arithmetic errors. Second, in derived systems a summary that folds the disclosure together with other values goes with it under closure. Part of aggregation's small residue after cleanup is therefore collateral rather than evidence the deleter found, although cleanup also deletes more aggregation evidence than any other type (Table~\ref{tab:evdel}). LightMem also loses aggregate values at write time, with 92\% residual recovery before deletion, the lowest of any type--system pair. In two cells native recovery of aggregation slightly exceeds residual recovery, which may miss aggregated values in objects without an annotated evidence source. The hidden-risk share remains a lower bound on the share of residual successes missed by native recovery.

\section{Case Study Details}
\label{app:case}

This appendix gives the full record of the fact in Figure~\ref{fig:case}: its evidence and cuts, where similarity to the request ranks each evidence turn, what S2 deletes at each pool size, and how entity expansion could reach a complete cut.

\paragraph{The fact and its cuts.}
The fact is that Arden Advisory, the firm the user's company hired, belongs to Nina Shah, the wife of its CFO, Ravi Shah. It rests on four links, each with two evidence turns: \emph{hiring}, the Q3 contract with Arden (E1) and a note that Ravi, the CFO, hired Arden (E2); \emph{ownership}, Arden's filing listing Nina Shah as sole owner (E3) and Nina's engagement letter as Arden's owner (E4); \emph{relationship}, the offsite dinner Ravi and his wife Nina host (E5) and a table booked for their wedding anniversary (E6); \emph{identity}, the same note (E2) and a staff line naming Ravi Shah as CFO (E7). Recovery needs one evidence turn from every link, so the minimal cuts are $\{$E1, E2$\}$, $\{$E3, E4$\}$, $\{$E5, E6$\}$, and $\{$E2, E7$\}$, two turns each.

\paragraph{Similarity scatters the evidence.}
For the request ``Forget that the firm we hired belongs to our CFO's wife'', the evidence turns rank 3 (E1), 14 (E3), 22 (E2), 36 (E5), 45 (E6), 53 (E4), and 141 (E7), so the main-table pool of 20 holds only E1 and E3, one evidence turn from each of two links. They share entities with the request but almost no wording, while records above and between them share its words without supporting the fact: a booking at the Arden Hotel, an old vendor list, an advisory budget, an advisory call, a note about a Nina in HR, an address on Arden Street. The disclosure itself ranks seventh; S0 deletes it, and it is excluded from every pool.

\paragraph{S2 completes a cut only at fifty.}
With a budget of ten, S2 deletes E1 from a pool of 10 and E1 and E3 from a pool of 20; the pool holds at most one evidence turn per link, S2 deletes what it holds, the budget never binds, and the fact stays derivable. From a pool of 30, E2 is in the pool and never selected, although with E1 it would complete the hiring cut: it reads as a routine record of a hire and says nothing about ownership. At 40, E5 joins the deleted turns. At 50, E6 enters the pool and is deleted; the relationship link has lost both evidence turns, and the fact is no longer derivable. At 60, E4 is deleted as well, emptying the ownership link too. E7 enters no pool we ran. In miniature, this is why the budget curve of Figure~\ref{fig:analysis}b is flat while the pool curve of Figure~\ref{fig:analysis}a falls.

\paragraph{Following entities could reach a complete cut.}
Every evidence turn is one hop from an entity the request refers to: Ravi, the CFO, links to E2, E5, E6, and E7; Arden, the firm, to E1, E2, E3, and E4; Nina, the wife, to E3, E4, E5, and E6. In this example, entity expansion could place a complete cut within a ten-turn pool; its effectiveness as a deletion strategy remains untested.

\section{Additional Related Work}
\label{app:related}

Beyond the five memories evaluated here, other agent memories \citep{park2023generative,wang2023longmem,xu2025amem,rasmussen2025zep,gutierrez2024hipporag,zhu2026helamem,park2026kumiho} vary the same two properties (\S\ref{sec:design}). Long-term memory is also evaluated for recall \citep{xu2022goldfish,tan2025membench}; forgetting in model parameters is studied through unlearning and knowledge editing \citep{jang2023knowledge,maini2024tofu,shi2025muse,meng2022rome,zhong2023mquake}; and contextual-privacy benchmarks test whether models disclose information in the wrong context \citep{mireshghallah2024confaide,shao2024privacylens}.

\end{document}